\documentclass[10pt,journal,compsoc]{IEEEtran}
\ifCLASSOPTIONcompsoc
  \usepackage[compress]{cite}
\else
  \usepackage{cite}
\fi

\ifCLASSINFOpdf
\usepackage[pdftex]{graphicx}
\DeclareGraphicsExtensions{.pdf,.jpeg,.png}
\else
\usepackage[dvips]{graphicx}
\graphicspath{{../eps/}}
\DeclareGraphicsExtensions{.eps}
\fi

\usepackage{times}
\usepackage{array}
\usepackage{epsfig}
\usepackage{graphicx}
\usepackage{amsmath}
\usepackage{cases}
\usepackage{bm}
\usepackage{cases}
\usepackage{amssymb}
\usepackage{multirow}
\usepackage{algorithm}
\usepackage{algpseudocode}
\usepackage{amsthm}
\newtheorem{theorem}{Theorem}
\usepackage{color}
\usepackage{xcolor}
\usepackage{mathrsfs}
\usepackage{epstopdf}
\usepackage{float}
\usepackage{ragged2e}
\renewcommand{\justify}{\leftskip=0pt \rightskip=0pt plus 0cm}

\ifCLASSOPTIONcompsoc
\usepackage[caption=false,font=footnotesize,labelfont=sf,textfont=sf]{subfig}
\else
  \usepackage[caption=false,font=footnotesize]{subfig}
\fi

\usepackage{fixltx2e}
\let\MYorigsubfloat\subfloat
\renewcommand{\subfloat}[2][\relax]{\MYorigsubfloat[]{#2}}
\usepackage{url}

\begin{document}
%
\title{Probabilistically Aligned View-unaligned Clustering with Adaptive Template Selection}

\author{Wenhua~Dong,
        Xiao-Jun~Wu*,
        Zhenhua~Feng, \emph{Senior Member, IEEE},
        Sara~Atito,
        Muhammad~Awais,
        and~Josef~Kittler, \emph{Life Member, IEEE} 
\IEEEcompsocitemizethanks{
\IEEEcompsocthanksitem W.~Dong is with the School of Science, Jiangnan University, Wuxi 214122, China, (e-mail: wenhua\_dong\_jnu@jiangnan.edu.cn).
\IEEEcompsocthanksitem X.~Wu and Z. Feng are with the School of Artificial Intelligence and Computer Science, Jiangnan University, Wuxi 214122, China, (e-mail: \{wu\_xiaojun; fengzhenhua\}@jiangnan.edu.cn).
\IEEEcompsocthanksitem S. Atito, M. Awais, and J. Kittler are with the Centre for Vision, Speech and Signal Processing, University of Surrey, Guildford GU2 7XH, UK, (e-mail: \{sara.atito, muhammad.awais, j.kittler\}@surrey.ac.uk).
}
}

\IEEEtitleabstractindextext{%
\begin{abstract}
\justify{In most existing multi-view modeling scenarios, cross-view correspondence (CVC) between instances of the same target from different views, like paired image-text data, is a crucial prerequisite for effortlessly deriving a consistent representation. Nevertheless, this premise is frequently compromised in certain applications, where each view is organized and transmitted independently, resulting in the view-unaligned problem (VuP). Restoring CVC of unaligned multi-view data is a challenging and highly demanding task that has received limited attention from the research community. To tackle this practical challenge, we propose to integrate the permutation derivation procedure into the bipartite graph paradigm for view-unaligned clustering, termed Probabilistically Aligned View-unaligned Clustering with Adaptive Template Selection (PAVuC-ATS). Specifically, we learn consistent anchors and view-specific graphs by the bipartite graph, and derive permutations applied to the unaligned graphs by reformulating the alignment between two latent representations as a 2-step transition of a Markov chain with adaptive template selection, thereby achieving the probabilistic alignment. The convergence of the resultant optimization problem is validated both experimentally and theoretically. Extensive experiments on six benchmark datasets demonstrate the superiority of the proposed PAVuC-ATS over the baseline methods.}
\end{abstract}

}

\maketitle

\IEEEdisplaynontitleabstractindextext
\IEEEpeerreviewmaketitle

\IEEEraisesectionheading{\section{Introduction}\label{sec:introduction}}

\IEEEPARstart{T}{he} rapid development of information technology and the widespread application of sensors in various fields have led to an explosive growth of multi-view data. For instance, in autonomous vehicles, multi-sensor data sourced from cameras, LiDAR, and radars significantly enhances perception and decision-making abilities, resulting in safer and more efficient driving. Likewise, in healthcare, data collected from wearable devices facilitates a comprehensive analysis of an individual's health status, enabling tailored interventions for optimal well-being. These diverse data, collected from various sources and domains, naturally raise the problem of multi-view clustering (MVC)~\cite{bickel2004multi,zhang2018generalized,li2019reciprocal,li2020multiview,xia2022tensorized,yang2022robust,trosten2023effects,
long2024s2mvtc}. The purpose of MVC is to leverage multiple representations of the data to reveal their underlying structure and membership relationships. By incorporating the complementary information from multiple views, MVC provides a more comprehensive understanding of the underlying category structure, resulting in more precise clustering assignments compared to single-view clustering methods.

\begin{figure}[!h]
\begin{center}
   \includegraphics[trim={102mm 84mm 106mm 66mm},clip,width=1\linewidth]{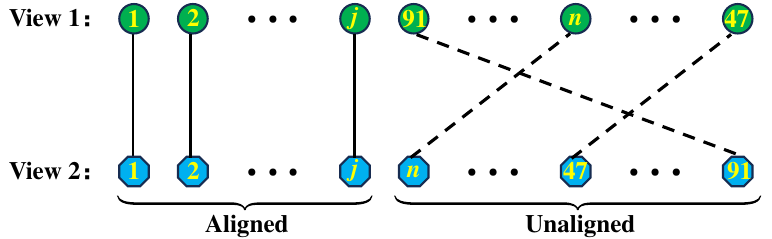}  
\end{center}
   \caption{The view-unaligned problem with an alignment ratio of $\rho\in[0,1]$, where digits and letters within circles and regular octagons represent the indices of instances, solid/dashed lines indicate known/unknown cross-view correspondences.}
\label{fig:1}
\end{figure}

In the past decade, MVC has garnered significant attention. Based on the extent of information available from multi-view data, existing multi-view clustering methods can be broadly categorized into three main types: (a) Complete and aligned multi-view clustering (CA-MVC), which assumes that there are no missing or unaligned instances in the multi-view data. Most current methods fall under the CA-MVC category. From the perspective of involved mathematical principles, these methods can be roughly divided into the following four categories, including multi-kernel learning~\cite{liu2023contrastive,wu2024low}, subspace learning~\cite{chen2023fast,mi2024fast}, graph~\cite{zhong2023self,dornaika2024towards}, and non-negative matrix-factorization~\cite{li2023robust,dong2024centric} based methods. (b) Incomplete multi-view clustering (IMVC)~\cite{liu2024sample,wan2024fast}, which supposes that some instances from certain views are missing. IMVC clusters the incomplete multi-view data by restoring missing instances. (c) View-unaligned clustering (VuC)~\cite{zhang2015constrained,wen2023unpaired}, which presumes that some instances of the same target from different views are unaligned, as illustrated in Figure~\ref{fig:1}, where the alignment ratio $\rho\in[0,1]$ characterizes the alignment level of the multi-view data, defined as the ratio of aligned samples to the total number of samples. VuC separates the unaligned multi-view data by recovering the cross-view correspondences of the unaligned instances. In this paper, we focus on the VuC with arbitrary alignment of $\rho\in[0,1]$.
\begin{figure*}[!t]
\begin{center}
\includegraphics[trim={34mm 57mm 38mm 48mm},clip,width=1\linewidth]{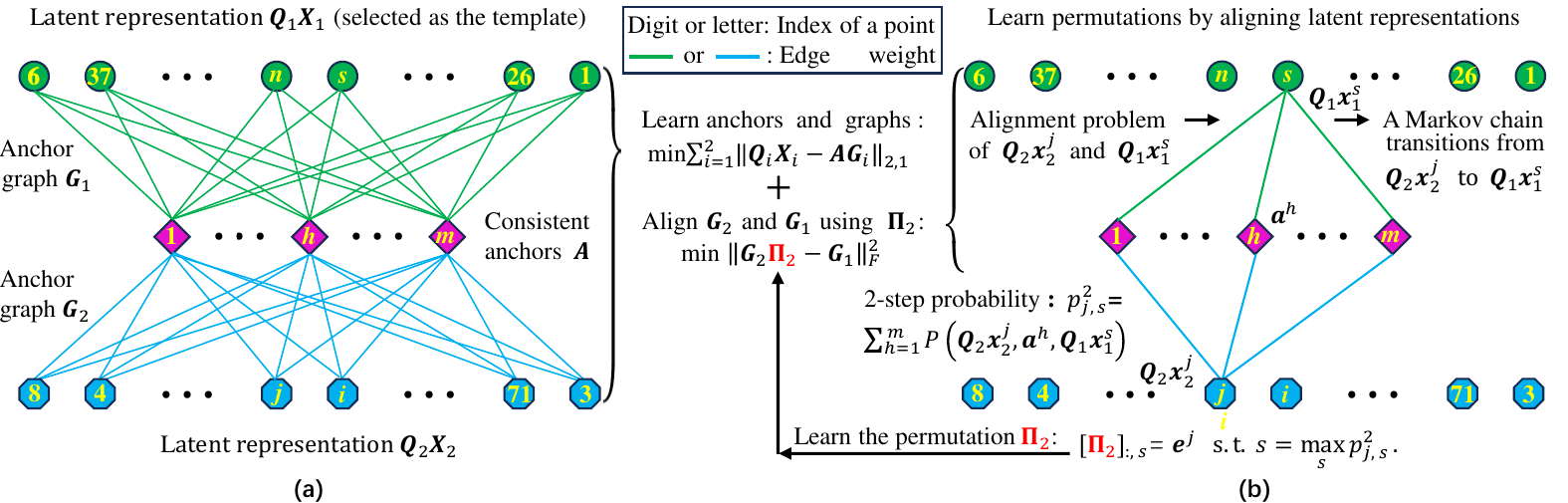}  
\end{center}
\caption{The PAVuC-ATS framework, as depicted in (a), initially learns anchors and graphs. Subsequently, it aligns $\bm{G}_{2}$ and $\bm{G}_{1}$ through the permutation $\bm{\Pi}_{2}$. To determine the permutation, the alignment between the latent representations $\bm{Q}_{2}\bm{x}_{2}^{j}$ and $\bm{Q}_{1}\bm{x}_{1}^{s}$ is reformulated as a Markov chain that transitions from state $\bm{Q}_{2}\bm{x}_{2}^{j}$ to state $\bm{Q}_{1}\bm{x}_{1}^{s}$ as illustrated in (b), achieving the probabilistic alignment, where $\bm{e}^{j}$ is the $j$-th orthogonal base vector.}
\label{fig:2}
\end{figure*}

The fundamental principle of MVC revolves around capturing the inherent similarity among data points utilizing diverse techniques, with self-representation and graph construction emerging as the most widely used methodologies. Self-representation based methods~\cite{brbic2018multi,zhang2023enhanced} represent data points using themselves as reference points, with typical examples including low-rank representation~\cite{liu2012robust} and sparse representation~\cite{elhamifar2013sparse}. However, the high computational complexity $\mathcal{O}(n^{3})$ for calculating an $n\times n$ coefficient matrix hinders the effective application of this technique to processing large-scale datasets, where $n$ is the number of samples. Different from the self-representation technique, the initial graph construction based approaches leverage graph structures to represent relationships among data points, such as spectral clustering~\cite{ng2001spectral}. Unfortunately, they encounter the selection of the appropriate graph construction approaches and high computational complexity $\mathcal{O}(n^{2})$. As an effective way to alleviate the aforementioned limitations, bipartite graph-based methods~\cite{li2015large,li2020multiview,xia2022tensorized} have garnered considerable interest from the research community. Instead of constructing an $n\times n$ similarity matrix, these methods aim to create an $m\times n$ ($m\ll n$) similarity graph by connecting $n$ data points to $m$ representative points (anchors), which models and analyzes complex relationships of the original data in a structured manner, thereby reducing the computational complexity from $\mathcal{O}(n^{2})$ to $\mathcal{O}(mn)$.

From the perspective of anchor selection strategies, existing bipartite graph-based methods can be primarily divided into two categories: (a) manually selecting anchors through strategies such as random sampling and $k$-means, followed by the construction of graphs, and (b) jointly learning both anchors and graphs by optimization. For instance, Li et al.~\cite{li2015large} proposed to use $k$-means on concatenated features as a means to select anchors, subsequently constructing anchor graphs leveraging the Gaussian kernel function. However, its performance heavily relies on the selected anchors and the similarity metric used for the data. To mitigate the performance fluctuations stemming from anchors selection via random sampling or $k$-means, Li et al.~\cite{li2020multiview} proposed a directly alternate sampling (DAS) strategy for selecting anchors that cover the point cloud of the data. In contrast, Xia et al.~\cite{xia2022tensorized} introduced a variance-based de-correlation anchor (VDA) selection strategy, ensuring that the selected anchors cover the whole categories of the data. Alternatively, various joint optimization techniques~\cite{kang2020large,wang2021fast,wang2022align,zhang2023let,zhang2024learning,10440580,liu2024learn} have been used to refine anchors and graphs, resulting in enhanced performance. Despite these methods achieve the impressive performance, they cannot effectively address the VuP due to their reliance on cross-view correspondences.

In many real-world applications, the assumption of CVC for multi-view data is often violated. For example, in the freeway monitoring system, cameras positioned along different road sections capture various views of the same target. However, due to timing differences, the VuP often occurs. Likewise, in medical diagnosis, doctors prescribe various tests for patients, including blood tests, X-rays, and MRI, which are regarded as different views. Nonetheless, variations in timing result in the VuP. Although the VuP is an issue that urgently needs to be solved in practical applications, it is seldom touched upon.

Recently, several studies have been conducted to address the partially view-unaligned problem (PVuP), i.e., $\rho\in(0,1]$. For instance, Huang et al.~\cite{huang2020partially} proposed to cluster the partially view-aligned data (PVC) by utilizing a differentiable surrogate of the Hungarian algorithm~\cite{kuhn1955hungarian,munkres1957algorithms}. However, the instance-level alignment achieved by PVC restricts its scalability. To alleviate this problem, Yang et al.~\cite{yang2021partially} reformulated the alignment problem as an identification task, resulting in the category-level alignment and enhanced scalability. Despite the encouraging performance obtained by the above approaches, they cannot tackle the fully view-unaligned problem (FVuP), i.e., $\rho=0$, due to their reliance on the partially aligned data to learn permutations or construct positive pairs. To address the problem, Wen et al.~\cite{wen2023unpaired} proposed a two-stage clustering solution for the VuP. This method first learns graphs using a graph clustering approach, and then aligns them by leveraging the graph structure matching mechanism. Nevertheless, the two-stage scheme could lead to a suboptimal solution and encounters high computational complexity, scaling up to $\mathcal{O}(n^{3})$ when $\rho=0$, during the graph structure matching stage.

To overcome the above limitations, we integrate the probabilistic alignment mechanism into the bipartite graph paradigm. Specifically, we initially learn consistent anchors and view-specific graphs using the bipartite graph, as depicted in Fig.~\ref{fig:2}(a), and then align the unaligned graphs with an adaptively selected template through permutations. Notably, to determine these permutations, as illustrated in Fig.~\ref{fig:2}(b), we reformulate the alignment between two latent representations as a 2-step transition of a Markov chain, thereby achieving the probabilistic alignment, where each latent representation or anchor is regarded as a state of a Markov chain, and the edge weights in the bipartite graph are treated as transition probabilities. Additionally, to alleviate the effect of the noisy template, we project each view to a latent space spanned by cross-view anchors, and use the~$\ell_{2,1}$ norm to characterize outliers in the data. In summary, the main innovations and contributions of the proposed method include:
\begin{itemize}
    \item We propose a probabilistically aligned clustering solution for the VuP with arbitrary alignment levels.
    \item The alignment between two latent representations is reformulated as a 2-step transition of a Markov chain with adaptive template selection.
    \item The integration of the bipartite graph and the probabilistic alignment mechanism guarantees efficiency and effectiveness of the proposed method.
    \item Extensive experiments on six benchmark datasets verify the superiority of PAVuC-ATS over twelve baseline approaches.
\end{itemize}

The remainder of this paper is organized as follows: Section~\ref{sec2} provides a brief overview of the preliminaries and related work. In Section~\ref{sec3}, we present a novel clustering solution designed to address the VuP. In Section~\ref{sec4}, extensive experiments conducted on six benchmark datasets demonstrate the advantages of the proposed method. Finally, the paper concludes in the last section.

\section{Preliminaries and Related Work}
\label{sec2}
\subsection{Preliminaries}
\textbf{Notation:} In this paper, we use bold capital letters to denote matrices, bold lowercase letters to represent vectors, and lowercase letters to signify scalars. Furthermore, we utilize square brackets with subscripts to denote individual entries within a matrix. For instance, $\bm{X}$ is a matrix, where $[\bm{X}]_{ij}$ denotes its $(i,j)$-th entry and, $[\bm{X}]_{i,:}$ and $[\bm{X}]_{:,j}$ denote its $i$-th row and $j$-th column, respectively. The horizontal and vertical concatenations of two matrices $\bm{X}_{1}$ and $\bm{X}_{2}$ are denoted by $[\bm{X}_{1},\bm{X}_{2}]$ and $[\bm{X}_{1};\bm{X}_{2}]$, respectively. The trace operator is represented as $Tr(\cdot)$ and, the Frobenius norm and $\ell_{2,1}$ norm are denoted by $\left|\cdot\right|_{F}$ and $\left|\cdot\right|_{2,1}$, respectively. Additionally, we use the symbols $\bm{I}$ and $\bm{1}$ to denote the identity matrix and all-ones vector, respectively. For clarity, Table~\ref{tab:1} summarizes the main notations used in this paper.
\begin{table}[!h]
\caption{The notations used in this paper.}
\label{tab:1}
\centering
\begin{tabular}{p{2cm}<{\centering}|p{6cm}}
\hline
\multirow{1}{*}{Notation} & \multirow{1}{*}{Description}\\
\hline
$\bm{X}_{i}$    & Feature matrix of the $i$-th view.\\
$\bm{Q}_{i}$    & Projection applied to $\bm{X}_{i}$.\\
$\bm{A}$        & Consistent anchors.\\
$\bm{G}_{i}$    & Anchor graph of $\bm{Q}_{i}\bm{X}_{i}$.\\
$\bm{\Pi}_{i}$  & Permutation applied to $\bm{G}_{i}$.\\
$\phi_{i}$      & The $i$-th weight factor.\\
\hline
$v$             & The number of views.\\
$k$             & The number of clusters.\\
$d_{i}$         & Feature dimension of $\bm{X}_{i}$.\\
$n$             & The number of samples.\\
$d_{l}$         & Dimension of the latent space.\\
$m$             & The number of anchors.\\
$\rho$          & Alignment ratio.\\
$\alpha$        & Control parameter.\\
$\mu$           & Trade-off parameter.\\
\hline
\end{tabular}
\end{table}

\noindent\textbf{Markov chain~\cite{serfozo2009basics}:} Suppose $\Omega$ is a countable set. A random process $\xi=\{\xi_{n}, n\geq 0\}$ on $\Omega$ is a Markov chain if, for $i,j\in\Omega$,
\begin{numcases}{}
P(\xi_{n+1}=j|\xi_{n}=i)=p_{i,j},\label{eq:1}\\
\sum_{j\in\Omega}p_{i,j}=1,\label{eq:2}\\                                                                                         
P(\xi_{n+1}=j|\xi_{0},\cdots,\xi_{n})=P(\xi_{n+1}=j|\xi_{n}), \label{eq:3}
\end{numcases}
where $p_{i,j}$ denotes the transition probability of a Markov chain jumping from state $i$ to state $j$. Based on the above definition, the $n$-step probability $p_{i_{0},i_{n}}^{n}$ of a Markov chain up to state $i_{n}$ can be calculated by~\cite{serfozo2009basics}:
\begin{equation}\label{eq:4}
\begin{aligned}
p_{i_{0},i_{n}}^{n}&\triangleq \sum_{(i_{0},i_{1},\cdots,i_{n})\in\mathcal{S}} P(\xi_{0}=i_{0},\cdots,\xi_{n}=i_{n})\\
&= \sum_{(i_{0},i_{1},\cdots,i_{n})\in\mathcal{S}} p_{0}p_{i_{0},i_{1}}\cdots p_{i_{n-1},i_{n}},                                  
\end{aligned}
\end{equation}
where $p_{0}=P(\xi_{0}=i_{0})$ is the probability of the initial state $i_{0}$ , $(i_{0},i_{1},\cdots,i_{n})$ denotes a sample path from state $i_{0}$ to state $i_{n}$, and $\mathcal{S}$ is a set composed of sample paths from state $i_{0}$ to state $i_{n}$.
\subsection{Related Work}
In this section, we revisit bipartite graph-based and view-unaligned clustering methods.

Consider a fully aligned data collection $\{\bm{X}_{i}\in\mathbb{R}^{d_{i}\times n}\}_{i=1}^{v}$ from $v$ views, where $\bm{X}_{i}$ represents the feature matrix of the $i$-th view comprising $n$ observations with dimension $d_{i}$.

Bipartite graph $\mathcal{G}(\mathcal{V},\mathcal{E})$ is a specialized form of graph whose vertices $\mathcal{V}$ are divided into two distinct and independent subsets $\mathcal{V}_{1}$ and $\mathcal{V}_{2}$, with all edges in the edge set $\mathcal{E}$ connecting vertices from $\mathcal{V}_{1}$ to those in $\mathcal{V}_{2}$. Most existing bipartite graph-based methods regard the data points from each view as vertices ($\mathcal{V}_{1}$), and seek $m$ ($m\ll n$) representative points (anchors) for the other set of vertices ($\mathcal{V}_{2}$), resulting in an $m\times n$ edge weighting matrix, referred to as an anchor graph. The general framework of the bipartite graph-based methods can be formulated as follows:
\begin{equation}\label{eq:5}
\begin{aligned}
&\underset{\bm{A}_{i},\bm{G}_{i}}{\mathrm{min}}\ \sum_{i=1}^{v}\mathcal{L}(\bm{X}_{i},\bm{A}_{i}\bm{G}_{i})
+\mu\ \psi(\bm{G}_{i})\\
&\mathrm{s.t.}\ \bm{G}_{i}\ge 0, \bm{G}_{i}^{T}\bm{1}\!=\!\bm{1},                                                                
\end{aligned}
\end{equation}
where $\mu>0$ is a trade-off parameter, $\bm{A}_{i}$ and $\bm{G}_{i}$ denote the anchors and graph from the $i$-th view separately, $\mathcal{L}$ is the loss function, and $\psi$ represents the specific regularization strategy. Under this framework, Kang et al.~\cite{kang2020large} proposed to simultaneously learn the anchors and graph for each view, enjoying high scalability. However, these anchors learned from different views may be unaligned, resulting in inaccurate graph fusion that negatively impacts the clustering performance. To address this issue, Wang et al.~\cite{wang2022align} put forward a strategy that first aligns anchors learned from each view and then fuses the aligned graphs, leading to improved performance. To refine the anchors derived from various views, Liu et al.~\cite{liu2024learn} proposed a anchor enhancement strategy that leverages view correlation. And Zhang et al.~\cite{zhang2023let} introduced a diverse anchor learning strategy aimed at learning different numbers of anchors and varying the dimensions of graphs from each view. Additionally, to capture the cross-view consensus and filter out view-specific noise, Liu et al.~\cite{10440580} proposed to learn a consistent anchor graph and a view-specific component from each view simultaneously. Although these methods achieve superior performance, they cannot effectively handle the VuP due to the unavailability of cross-view correspondences.

Recently, several methods~\cite{huang2020partially,yang2021partially} have been proposed to address the PVuP by integrating data alignment and representation learning within a single network framework. However, these approaches rely on the data being partially aligned to learn the permutations or construct positive pairs, hindering them from addressing the FVuP. To overcome the limitation, Wen et al.~\cite{wen2023unpaired} proposed a two-stage method for VuC, which involves first learning graph matrices from different views utilizing a graph clustering approach, followed by their alignment, leveraging the graph structure matching mechanism. The objective function can be formulated as follows:
\begin{equation}\label{eq:6}
\begin{aligned}
\underset{\bm{\Pi}_{i}}{\mathrm{min}}\ \sum_{i\neq t}\left\|\bm{\Pi}_{i}^{T}\bm{S}_{i}\bm{\Pi}_{i}-\bm{S}_{t}\right\|_{F}^{2},      
\end{aligned}
\end{equation}
where $\bm{S}_{i}$ represents the graph matrix learned from the $i$-th view, $\bm{S}_{t}$ denotes the selected alignment template, and $\bm{\Pi}_{i}$ ($i\neq t$) represents the permutation applied to $\bm{S}_{i}$ that satisfies the properties: $\bm{\Pi}_{i}\bm{1}=\bm{1}, \bm{\Pi}_{i}^{T}\bm{1}=\bm{1}$, with $[\bm{\Pi}_{i}]_{hj}\in\{0,1\}$. The above objective function can be solved using the projected fixed-point algorithm~\cite{lu2016fast} with the computational complexity $\mathcal{O}(n^{3})$ in the case of $\rho=0$.

\section{The Proposed Method}
\label{sec3}
In this section, we present a novel MVC method for addressing the VuP with an alignment ratio of $\rho\in[0,1]$. We also provide the complexity and convergence analysis of the resultant optimization problem.
\subsection{Model Formulation}
Given a multi-view data set $\{\bm{X}_{i}\in\mathbb{R}^{d_{i}\times n}\}_{i=1}^{v}$ with an alignment ratio of $\rho\in[0,1]$. Without loss of generality, the view-unaligned setting is represented as $\bm{X}_{i}=[\bm{X}_{i}^{a},\bm{X}_{i}^{u}]$, $i=1,2,\cdots,v$, where $\bm{X}_{i}^{a}$ /$\bm{X}_{i}^{u}$ denote the aligned/unaligned observations. We assume that the $t$-th view is selected as the alignment template. With an abuse of notation, the corresponding latent representation and anchor graph are also referred to as templates.

To address the VuP, we confront three pivotal challenges: (a) devising an efficient and effective alignment mechanism for the unaligned data; (b) enhancing the algorithm's scalability; and (c) adaptively selecting the optimal alignment template. To handle the first challenge, we recast the alignment between two latent representations as a 2-step transition of a Markov chain, resulting in the probabilistic alignment mechanism, with the computational complexity of $\mathcal{O}(n^{2})$ when $\rho=0$. To tackle the second challenge, we incorporate the probabilistic alignment mechanism into the bipartite graph framework, thereby enjoying high scalability. Regarding the last challenge, we formulate a data-driven strategy for choosing the most appropriate alignment template. Consequently, the objective function can be formulated as follows:
\begin{equation}\label{eq:7}
\begin{aligned}
&\underset{\Delta}{\mathrm{min}}
\sum_{i=1}^{v}(\phi_{i})^{\alpha}\left\|\bm{Q}_{i}\bm{X}_{i}\!-\!\bm{A}\bm{G}_{i}\right\|_{2,1}
\!+\!\mu\sum_{i\neq t}\left\|\bm{G}_{i}\bm{\Pi}_{i}\!-\!\bm{G}_{t}\right\|_{F}^{2}\\                                                
&\mathrm{s.t.}\ \bm{Q}_{i}\bm{Q}_{i}^{T}\!=\!\bm{I}, \bm{A}^{T}\bm{A}\!=\!\bm{I}, \bm{G}_{i}\ge 0, \bm{G}_{i}^{T}\bm{1}\!=\!\bm{1}, \sum_{i=1}^{v}\phi_{i}\!=\!1,
\end{aligned}
\end{equation}
where $\Delta=\{\bm{Q}_{i},\bm{A},\bm{G}_{i},\bm{\Pi}_{i},\phi_{i}\}_{i=1}^{v}$ is a set comprised of variables to be optimized, $\phi_{i}$ is the weighting factor corresponding to the $i$-th view, indicating its importance among all views, the parameter $\alpha>1$ controls the distribution of weights, and $\bm{G}_{t}$ represents the alignment template. Furthermore, $\bm{\Pi}_{i}$ denotes the permutation applied to $\bm{G}_{i}$. If $i=t$, then $\bm{\Pi}_{i}=\bm{I}$; otherwise, $\bm{\Pi}_{i}$ is a block diagonal matrix, i.e., $\begin{bmatrix}\bm{I} & 0 \\0 & \bm{\Pi}_{i}^{u}\end{bmatrix}$, where $\bm{\Pi}_{i}^{u}$ is a permutation matrix used to permutate the unaligned subgraph within $\bm{G}_{i}$. The block diagonal structure of $\bm{\Pi}_{i}$ ($i\neq t$) enables us to address the VuP with arbitrary alignment levels. In Eq.~(\ref{eq:7}), the first term is utilized to learn cross-view anchors and view-specific graphs by the bipartite graph, while the second term is employed to align these graphs with the adaptively selected template through permutations, simultaneously fostering consensus. The objective of learning cross-view anchors, which serve as bases, is to establish a unified benchmark for anchor graph learning from different views, thereby facilitating the subsequent alignment of latent representations. Furthermore, to mitigate the impact of the noise template, we project each view to a latent space spanned by cross-view anchors, and use the~$\ell_{2,1}$ norm to characterize outliers in the data.

\subsection{Optimization Algorithm}
The objective function in Eq.~(\ref{eq:7}) is not jointly convex with respect to all variables, posing a challenge for direct optimization. Hence, we adopt an alternative rule in which we update one variable while keeping the others fixed.

To derive the optimization algorithm, we first utilize the properties of the $\ell_{2,1}$ norm to rewrite the first term in Eq.~(\ref{eq:7}) as $Tr(\bm{E}_{i}\bm{\Lambda}_{i}\bm{E}_{i}^{T})$, where $\bm{E}_{i}=\bm{Q}_{i}\bm{X}_{i}-\bm{A}\bm{G}_{i}$, and $\bm{\Lambda}_{i}$ is a diagonal matrix whose $j$-th diagonal entry is defined as follows:
\begin{equation}\label{eq:8}
\begin{aligned}
{[\bm{\Lambda}_{i}]}_{jj} = \frac{1}{2\left\|[\bm{Q}_{i}\bm{X}_{i}]_{:,j}-[\bm{A}\bm{G}_{i}]_{:,j}\right\|_{2}}.                  
\end{aligned}
\end{equation}

\textbf{Solving} $\bm{Q}_{i}$ \textbf{with other variables fixed}. The sub-problem with respect to $\bm{Q}_{i}$ is presented below:
\begin{equation}\label{eq:9}
\begin{aligned}
\underset{\bm{Q}_{i}}{\mathrm{min}}\ Tr(\bm{E}_{i}\bm{\Lambda}_{i}\bm{E}_{i}^{T})
\ \ \mathrm{s.t.}\ \bm{Q}_{i}\bm{Q}_{i}^{T}=\bm{I},                                                                               
\end{aligned}
\end{equation}
which is challenging to solve directly due to the constraint $\bm{Q}_{i}\bm{Q}_{i}^{T}$ $=\bm{I}$. Therefore, we relax it to $\bm{Q}_{i}^{T}\bm{Q}_{i}\!=\!\bm{Q}_{i}\bm{Q}_{i}^{T}\!=\!\bm{I}$ according to the work in~\cite{zhang2018generalized}, leading to the following optimization problem:
\begin{equation}\label{eq:10}
\begin{aligned}
\underset{\bm{Q}_{i}}{\mathrm{max}}\ Tr(\bm{Q}_{i}\bm{B}_{i})                                                                    
\ \ \mathrm{s.t.}\ \bm{Q}_{i}^{T}\bm{Q}_{i}=\bm{Q}_{i}\bm{Q}_{i}^{T}=\bm{I},
\end{aligned}
\end{equation}
where $\bm{B}_{i}=\bm{X}_{i}\bm{\Lambda}_{i}\bm{G}_{i}^{T}\bm{A}^{T}$. We solve the above optimization problem using Theorem~\ref{thm:1}:
\begin{theorem}[{\cite{huang2013spectral}}]\label{thm:1}
Given the optimization problem with respect to $\bm{Y}$: $\mathrm{max}\ Tr(\bm{Y}^{T}\bm{X})$ $\mathrm{s.t.}\ \bm{Y}^{T}\bm{Y} =\bm{Y}\bm{Y}^{T}=\bm{I}$, the optimal solution of $\bm{Y}$ is given by $\bm{U}\bm{V}^{T}$, where $\bm{U}\bm{\Sigma}\bm{V}^{T}$ is the singular value decomposition (SVD) of $\bm{X}$.
\end{theorem}

Following Theorem 1, we can get the optimal solution of $\bm{Q}_{i}^{T}$ by $\bm{U}\bm{V}^{T}$, where $\bm{U}\bm{\Sigma}\bm{V}^{T}$ is the SVD of $\bm{B}_{i}$.

\textbf{Solving} $\bm{A}$ \textbf{with other variables fixed}. The sub-problem with respect to $\bm{A}$ is presented below:
\begin{equation}\label{eq:11}
\begin{aligned}
\underset{\bm{A}}{\mathrm{min}}\sum_{i=1}^{v}(\phi_{i})^{\alpha}\ Tr(\bm{E}_{i}\bm{\Lambda}_{i}\bm{E}_{i}^{T})                    
\ \ \mathrm{s.t.}\ \bm{A}^{T}\bm{A}=\bm{I}.
\end{aligned}
\end{equation}
By expanding the above objective function and discarding the terms unrelated to $\bm{A}$, we have:
\begin{equation}\label{eq:12}
\begin{aligned}
\underset{\bm{A}}{\mathrm{max}}\ Tr(\bm{A}\bm{C})                                                                                 
\ \ \mathrm{s.t.}\ \bm{A}^{T}\bm{A}=\bm{I},
\end{aligned}
\end{equation}
where $\bm{C}=\sum_{i=1}^{v}(\phi_{i})^{\alpha}\bm{G}_{i}\bm{\Lambda}_{i}\bm{X}_{i}^{T}\bm{Q}_{i}^{T}$. Similar to solving the $\bm{Q}_{i}$ sub-problem, we can get the optimal solution of $\bm{A}$ by $\bm{U}\bm{V}^{T}$, where $\bm{U}\bm{\Sigma}\bm{V}^{T}$ is the SVD of $\bm{C}$.

\textbf{Solving} $\bm{G}_{i}$ \textbf{with other variables fixed}. The sub-problem with respect to $\bm{G}_{i}$ ($i\ne t$) is presented below:
\begin{equation}\label{eq:13}
\begin{aligned}
&\underset{\bm{G}_{i}} {\mathrm{min}}\ (\phi_{i})^{\alpha}\ Tr(\bm{E}_{i}\bm{\Lambda}_{i}\bm{E}_{i}^{T})                    +\mu\left\|\bm{G}_{i}\bm{\Pi}_{i}-\bm{G}_{t}\right\|_{F}^{2}\\                                                                   
&\mathrm{s.t.}\ \bm{G}_{i}\ge 0, \bm{G}_{i}^{T}\bm{1}=\bm{1},
\end{aligned}
\end{equation}
which can be represented column-wisely as $n$ projection capped simplex problems, as defined in~\cite{wang2015projection}. For $j>0$, we have:
\begin{equation}\label{eq:14}
\begin{aligned}
\underset{\bm{g}_{i}^{j}} {\mathrm{min}}\ \left\|\bm{g}_{i}^{j}-\bm{h}_{i}^{j}\right\|^{2}                                       
\ \ \mathrm{s.t.}\ \bm{g}_{i}^{j}\ge 0, (\bm{g}_{i}^{j})^{T}\bm{1}=1,
\end{aligned}
\end{equation}
where $\bm{g}_{i}^{j}$ denotes the $j$-th column of the matrix $\bm{G}_{i}$, $\bm{h}_{i}^{j}=\frac{\gamma_{i}^{j}\bm{A}^{T}\bm{Q}_{i}\bm{x}_{i}^{j}+\mu\bar{\bm{g}}_{i}^{j}}{\gamma_{i}^{j}+\mu}$, $\gamma_{i}^{j}=(\phi_{i})^{\alpha}[\bm{\Lambda}_{i}]_{jj}$, and $\bar{\bm{g}}_{i}^{j}$ is the $j$-th column of $\bm{G}_{t}\bm{\Pi}_{i}^{T}$. The Lagrangian is:
\begin{equation}\label{eq:15}
\begin{aligned}
\left\|\bm{g}_{i}^{j}-\bm{h}_{i}^{j}\right\|^{2}-\eta_{i}^{j}((\bm{g}_{i}^{j})^{T}\bm{1}-1)-(\bm{\delta}_{i}^{j})^{T}\bm{g}_{i}^{j}, 
\end{aligned}
\end{equation}
where $\eta_{i}^{j}$ and $\bm{\delta}_{i}^{j}$ are the Lagrange multipliers. At the optimal solution of $\bm{g}_{i}^{j}$, the following KKT conditions hold:
\begin{equation}\label{eq:16}
\begin{aligned}
\left\{\begin{matrix}
\bm{g}_{i}^{j}-\bm{h}_{i}^{j}-\eta_{i}^{j}\bm{1}-\bm{\delta}_{i}^{j}=0,~~\\
(\bm{g}_{i}^{j})^{T}\bm{1}=1,~~~~~~~~~~~~~~~~~~~~~~~\\                                                                          
\bm{\delta}_{i}^{j}\odot\bm{g}_{i}^{j}=0,~~~~~~~~~~~~~~~~~~~~~\\
\end{matrix}\right.
\end{aligned}
\end{equation}
where $\odot$ denotes the Hadamard product operator. The system of equations in (\ref{eq:16}) can be solved by:
\begin{equation}\label{eq:17}
\begin{aligned}
\bm{g}_{i}^{j}=\mathrm{max}(\bm{h}_{i}^{j}+\eta_{i}^{j}\bm{1},0),\ \                                                            
\eta_{i}^{j} = \frac{1-(\bm{h}_{i}^{j})^{T}\bm{1}}{m},
\end{aligned}
\end{equation}
where $m$ is the number of anchors.

Similarly, we can find the optimal solution of $\bm{G}_{t}$ column-wisely by:
\begin{equation}\label{eq:18}
\begin{aligned}
\bm{g}_{t}^{j}=\mathrm{max}(\tilde{\bm{h}}_{t}^{j}+\eta_{t}^{j}\bm{1},0),\ \                                                    
\eta_{t}^{j} = \frac{1-(\tilde{\bm{h}}_{t}^{j})^{T}\bm{1}}{m},
\end{aligned}
\end{equation}
where $\tilde{\bm{h}}_{t}^{j}=\frac{\gamma_{t}^{j}\bm{A}^{T}\bm{Q}_{t}\bm{x}_{t}^{j}+\mu\tilde{\bm{g}}_{t}^{j}}{\gamma_{t}^{j}+\mu(v-1)}$, and $\tilde{\bm{g}}_{t}^{j}$ is the $j$-th column of $\sum_{i\ne t}\bm{G}_{i}\bm{\Pi}_{i}$.

\textbf{Solving} $\bm{\Lambda}_{i}$ \textbf{with other variables fixed}. The solution of $\bm{\Lambda}_{i}$ is given by Eq.~(\ref{eq:8}).

\textbf{Solving} $\bm{\Pi}_{i}$ ($i\neq t$) \textbf{with other variables fixed}. The sub-problem with respect to $\bm{\Pi}_{i}$ ($i\neq t$) is presented below:
\begin{equation}\label{eq:19}
\begin{aligned}
&\underset{\bm{\Pi}_{i}}{\mathrm{min}}\ \left\|\bm{G}_{i}\bm{\Pi}_{i}-\bm{G}_{t}\right\|_{F}^{2}\\                              
&\mathrm{s.t.}\ \bm{\Pi}_{i}\bm{1}=\bm{1}, {\bm{\Pi}_{i}}^{T}\bm{1}=\bm{1}, [\bm{\Pi}_{i}]_{hj}\in\{0,1\}.
\end{aligned}
\end{equation}
Under the given constraints, the above optimization problem is equivalent to:
\begin{equation}\label{eq:20}
\begin{aligned}
&\underset{\bm{\Pi}_{i}}{\mathrm{max}}\ Tr(\bm{\Pi}_{i}^{T}\bm{G}_{i}^{T}\bm{G}_{t})\\                                          
&\!=\!\underset{(h_{1},h_{2},\cdots,h_{n})}{\mathrm{max}}\ \!(\bm{g}_{i}^{h_{1}})^{T}\bm{g}_{t}^{1}\!+\!(\bm{g}_{i}^{h_{2}})^{T}\bm{g}_{t}^{2}\!+\!\cdots\!+\!(\bm{g}_{i}^{h_{n}})^{T}\bm{g}_{t}^{n},
\end{aligned}
\end{equation}
where $(h_{1},h_{2},\cdots,h_{n})$ is a permutation of the sequence $(1,2,$ $\cdots,n)$. To solve the optimization problem in Eq.~(\ref{eq:20}), taking two views, $\bm{X}_{1}$ and $\bm{X}_{2}$, as a showcase, we reformulate the alignment between $\bm{Q}_{2}\bm{x}_{2}^{j}$ and $\bm{Q}_{1}\bm{x}_{1}^{s}$ as a 2-step transition of a Markov chain that transitions from state $\bm{Q}_{2}\bm{x}_{2}^{j}$ to state $\bm{Q}_{1}\bm{x}_{1}^{s}$, as illustrated in Fig.~\ref{fig:2}(b). Notably, the specified constraints on $\bm{G}_{i}$ in Eq.~(\ref{eq:7}): $\bm{G}_{i}\ge 0, \bm{G}_{i}^{T}\bm{1}\!=\!\bm{1}$, $i=1,2,\cdots, v$ allow us to address this alignment issue from the perspective of Markov chains. In this context, each latent representation (or anchor) is treated as a state of a Markov chain, while the edge weight between the latent representation and the anchor in the bipartite graph is regarded as the transition probability. Fig.~\ref{fig:2}(b) indicates that there are $m$ sample paths from state $\bm{Q}_{2}\bm{x}_{2}^{j}$ to state $\bm{Q}_{1}\bm{x}_{1}^{s}$. Therefore, the 2-step probability can be calculated by:
\begin{equation}\label{eq:21}
\begin{aligned}
p_{j,s}^{2} &= \sum_{h=1}^{m}P(\bm{Q}_{2}\bm{x}_{2}^{j},\bm{a}^{h},\bm{Q}_{1}\bm{x}_{1}^{s})\\
& = \sum_{h=1}^{m}P(\bm{Q}_{2}\bm{x}_{2}^{j})P(\bm{a}^{h}|\bm{Q}_{2}\bm{x}_{2}^{j})P(\bm{Q}_{1}\bm{x}_{1}^{s}|\bm{a}^{h})\\      
& = \sum_{h=1}^{m}P(\bm{Q}_{2}\bm{x}_{2}^{j})[\bm{G}_{2}]_{hj}[\bm{G}_{1}]_{hs}\\
& = P(\bm{Q}_{2}\bm{x}_{2}^{j})(\bm{g}_{2}^{j})^{T}\bm{g}_{1}^{s}.
\end{aligned}
\end{equation}
where the second equation in~(\ref{eq:21}) holds due to the condition~(\ref{eq:3}) specified in the definition of a Markov chain. To ascertain the probability of the initial state $\bm{Q}_{2}\bm{x}_{2}^{j}$ in Eq.~(\ref{eq:21}), we leverage the discriminative information inherent within the data. Generally, a larger variance of a data point suggests that it carries more discriminative information. Consequently, a random particle is more likely to be found in an initial state corresponding to a data point with a larger variance. Therefore, the probability of the initial state $\bm{Q}_{2}\bm{x}_{2}^{j}$ can be calculated by:
\begin{equation}\label{eq:22}
\begin{aligned}
P(\bm{Q}_{2}\bm{x}_{2}^{j}) = \frac{\mathrm{Var}(\bm{x}_{2}^{j})}{\sum_{h=1}^{n}\mathrm{Var}(\bm{x}_{2}^{h})},                   
\end{aligned}
\end{equation}
where $\mathrm{Var}(\cdot)$ denotes the variance operator.

Thus, we can rewrite the optimization problem in Eq.~(\ref{eq:20}) column-wisely as follows, for $j>0$:
\begin{equation}\label{eq:23}
\begin{aligned}
{[\bm{\Pi}_{i}]}_{:,s} = \bm{e}_{j} \ \ \mathrm{s.t.}\ s=\underset{s}{\mathrm{max}}\ p_{j,s}^{2},                               
\end{aligned}
\end{equation}
which can be effectively solved by utilizing an exhaustive search among $n$ candidates $\{p_{j,s}^{2}\}_{s=1}^{n}$.

Although the initial state probability in Eq.~(\ref{eq:22}) remains constant during the alignment of the latent representation $\bm{Q}_{2}\bm{x}_{2}^{j}$, thus not affecting the solution of ${[\bm{\Pi}_{i}]}_{:,s}$, it provides valuable insights into the alignment order for latent representations. This information can therefore be leveraged to refine cross-view correspondences. Furthermore, to establish a 1-to-1 cross-view correspondence, we set the values of the $s$-th column in $\bm{G}_{i}^{T}\bm{G}_{t}$ to $-1$, after selecting the latent representation $\bm{Q}_{t}\bm{x}_{t}^{s}$ as the counterpart of $\bm{Q}_{i}\bm{x}_{i}^{j}$, to ensure that the $s$-th latent representation $\bm{Q}_{t}\bm{x}_{t}^{s}$ is not selected again in the subsequent alignment process.

\textbf{Solving} $\phi_{i}$ \textbf{with the other variables fixed}. The sub-problem with respect to $\phi_{i}$ is presented below:
\begin{equation}\label{eq:24}
\begin{aligned}
&\underset{\phi_{i}}{\mathrm{min}}\ (\phi_{i})^{\alpha}\varepsilon_{i}\ \ \mathrm{s.t.}\ \sum_{i=1}^{v}\phi_{i}\!=\!1,          
\end{aligned}
\end{equation}
where $\varepsilon_{i}=\left\|\bm{Q}_{i}\bm{X}_{i}-\bm{A}\bm{G}_{i}\right\|_{2,1}$. The Lagrangian is:
\begin{equation}\label{eq:25}
\begin{aligned}
\mathcal{L}(\phi_{i},\lambda)=(\phi_{i})^{\alpha}\varepsilon_{i}-\lambda(\sum_{i=1}^{v}\phi_{i}-1),                            
\end{aligned}
\end{equation}
where $\lambda$ is the Lagrange multiplier. Taking the partial derivatives of $\mathcal{L}(\phi_{i},\lambda)$ with respect to $\phi_{i}$ and $\lambda$ separately, and then setting them to 0, we have:
\begin{equation}\label{eq:26}
\begin{aligned}
\left\{\begin{matrix}
\phi_{i}=(\frac{\lambda}{\alpha \varepsilon_{i}})^{\frac{1}{\alpha-1}},~~\\
\sum_{i=1}^{v}\phi_{i}=1,~~~~~\\                                                                                                
\end{matrix}\right.
\end{aligned}
\end{equation}
which can be solved by:
\begin{equation}\label{eq:27}
\begin{aligned}
\phi_{i}=\frac{(\varepsilon_{i})^{\frac{1}{1-\alpha}}}{\sum_{i=1}^{v}(\varepsilon_{i})^{\frac{1}{1-\alpha}}}.                  
\end{aligned}
\end{equation}

Finally, we fuse the aligned anchor graphs learned from Eq.~(\ref{eq:7}) by calculating their average, i.e.,
\begin{equation}\label{eq:28}
\begin{aligned}
\bar{\bm{G}}=\frac{\sum_{i=1}^{v}\bm{G}_{i}\bm{\Pi}_{i}}{v}.                                                                  
\end{aligned}
\end{equation}
Subsequently, the rank-$k$ truncated SVD is applied to $\bar{\bm{G}}$, yielding $\bm{U}\bm{\Sigma}\bm{V}^{T}$. The left singular value vectors $\bm{U}$ are then utilized as input for $k$-means to obtain clustering assignments. The entire process is summarized in Algorithm~\ref{alg:1}.
\begin{algorithm}[h!]
\caption{: View-unaligned clustering by PAVuC-ATS}
\label{alg:1}
\begin{algorithmic}
\Require
Data matrices: $\{\bm{X}_{i}\}_{i=1}^{v}$, the number of clusters $k$, and\\
~~~~~~~~maxIter = 60.
\Ensure
Clustering assignments.\\
~1: \textbf{Initialization:}~Initialize $\bm{A}$ and $\{\bm{G}_{i}\}_{i=1}^{v}$ randomly,\\
~~~~ $\{\bm{\Pi}_{i}=\bm{I}\}_{i=1}^{v}$, $\{\bm{\Lambda}_{i}=\bm{I}\}_{i=1}^{v}$, and $\{\phi_{i}=\frac{1}{v}\}_{i=1}^{v}$.\\
~2:~\textbf{while} not converged \textbf{do}\\
~3:~~ Fix others and update $\bm{Q}_{i}$ via Eq.~(\ref{eq:10});\\
~4:~~ Fix others and update $\bm{A}$ via Eq.~(\ref{eq:12});\\
~5:~~ Fix others and update $\bm{G}_{i}$ via Eqs.~(\ref{eq:17}) and (\ref{eq:18});\\
~6:~~ Fix others and update $\bm{\Lambda}_{i}$ via Eq.~(\ref{eq:8});\\
~7:~~ Fix others and update $\bm{\Pi}_{i}$ via Eq.~(\ref{eq:23});\\
~8:~~ Fix others and update $\phi_{i}$ via Eq.~(\ref{eq:27});\\
~9:~~ Check convergence:~~~~~~~~~~~~~~~~~~~~~~~~~~~~~~~~~\\
~~~~~~ $(\mathrm{obj}(j\!-\!1)\!-\!\mathrm{obj}(j))/\mathrm{obj}(j)\!<\!10^{-7}$ or $j\!>\! \mathrm{maxIter}$.\\
10:~\textbf{end while}\\
11:~Compute the rank-$k$ truncated SVD of $\bar{\bm{G}}$, i.e., $\bm{U}\bm{\Sigma}\bm{V}^{T}$.\\
12:~Apply $k$-means to the left singular vectors $\bm{U}$.
\end{algorithmic}
\end{algorithm}
\subsection{Adaptive Template Selection}\label{sec:3.3}
In this section, we present an adaptive template selection strategy: allowing the data to determine the optimal template. In Eq.~(\ref{eq:7}), the weighting factor $\phi_{i}$, $i=1,2,\cdots, v$ indicates the relative importance of the $i$-th latent representation compared to the others~\cite{cai2013multi}. Given $\alpha>1$, Eq.~(\ref{eq:27}) reveals that $\phi_{i}$ is a monotonically decreasing function with respect to $\varepsilon_{i}$. Therefore, a higher value of $\phi_{i}$ signifies a lower reconstruction error for the $i$-th latent representation, making it more prominent among all latent representations. This insight provides a practical approach for selecting the optimal template, which can be formally formulated as: $t=\{i|\mathrm{max}_{i}\ \phi_{i}\}$, where $t$ denotes the index of the template.
\subsection{Complexity and Convergence Analysis}\label{sec:3.4}
We perform the complexity analysis on the data setting with an alignment ratio of $\rho=0$. Table~\ref{tab:2} indicates that the time and space complexities of Algorithm~\ref{alg:1} closely approximate $\mathcal{O}(\lambda_{1}n^{2})$ and $\mathcal{O}(\lambda_{2}n^{2})$ respectively, given that $d_{i},d_{l},m,r,v\ll n$, where $\lambda_{1}, \lambda_{2}>0$ are constants, and $r$ denotes the number of iterations. Despite the computational complexity of Algorithm~\ref{alg:1}, scaling linearly with $n^{2}$, its main computational overhead arises from the matrix multiplication during the update of $\bm{\Pi}_{i}$ ($i\neq t$), specifically the computations of $\bm{G}_{i}^{T}\bm{G}_{t}$ ($i\neq t$) in Eq.~(\ref{eq:21}). Additionally, to reduce time and space complexities, we vectorize the permutation matrix. For example, suppose $\bm{X}\in \mathbb{R}^{2\times 3}$ is a matrix, and $\bm{\Pi}=\begin{bmatrix}0&0&1;~1&0&0;~0&1&0\end{bmatrix}$ is a permutation applied to $\bm{X}$. By vectorizing $\bm{\Pi}$ into $\bm{\pi}={[2,3,1]}^{T}$, we can express the equation $\bm{X}\bm{\Pi}=\bm{X}(:,\bm{\pi})$, leading to a reduced time and space requirements.
\begin{table}[h!]
\caption{A summary of the complexity of Algorithm~\ref{alg:1}.}
\label{tab:2}
\centering
\begin{tabular}{p{2.0cm}<{\centering}p{3cm}<{\centering}p{2.5cm}<{\centering}}
\hline
\multicolumn{1}{c}{\multirow{1}{*}{Variable}} &\multirow{1}{*}{Time complexity} &\multirow{1}{*}{Space complexity}\\
\hline
\multirow{1}{*}{$\bm{Q}_{i}\in\mathbb{R}^{d_{l}\times d_{i}}$}&$\mathcal{O}(d_{i}d_{l}^{2}+d_{l}^{3}+d_{i}mn)$ &$\mathcal{O}(d_{i}d_{l})$\\
\multirow{1}{*}{$\bm{A}\in\mathbb{R}^{d_{l}\times m}$}        &$\mathcal{O}(md_{l}^{2}+d_{l}^{3}+d_{i}mn)$  &$\mathcal{O}(d_{l}m)$\\
\multirow{1}{*}{$\bm{G}_{i}\in\mathbb{R}^{m\times n}$}        &$\mathcal{O}(d_{i}d_{l}m+d_{i}mn)$           &$\mathcal{O}(mn)$\\
\multirow{1}{*}{$\bm{\Lambda}_{i}\in\mathbb{R}^{n\times n}$}  &$\mathcal{O}(d_{i}d_{l}n+d_{l}mn)$           &$\mathcal{O}(n^{2})$\\
\multirow{1}{*}{$\bm{\Pi}_{i}\in\mathbb{R}^{n\times n}$}      &$\mathcal{O}(mn^{2})$                        &$\mathcal{O}(n^{2})$ \\
\multirow{1}{*}{$\phi_{i}\in\mathbb{R}$}                      &$\mathcal{O}(d_{i}d_{l}n+d_{l}mn)$           &$\mathcal{O}(1)$\\
\hline
\multirow{1}{*}{Summation}                                    &$\mathcal{O}(\lambda_{1} n^{2})$      &$\mathcal{O}(\lambda_{2} n^{2})$\\
\hline
\end{tabular}
\end{table}

The optimization problem presented in Eq.~(\ref{eq:7}) is not jointly convex with respect to all variables. Nevertheless, the proposed alternative optimization rule ensures that Algorithm~\ref{alg:1} converges to a local minimum. We provide a theoretical proof to support this claim.

For ease of description, we rewrite the optimization problem presented in Eq.~(\ref{eq:7}) as:
\begin{equation}\label{eq:29}
\begin{split}
{\mathrm{min}}\ \mathcal{H}\Big(\{\bm{Q}_{i}, \bm{G}_{i}, \bm{\Lambda}_{i}, \bm{\Pi}_{i},\phi_{i}\}_{i=1}^{v},\bm{A}\Big). 
\end{split}
\end{equation}

During the $(j + 1)$-th iteration, we solve for one variable while keeping the remaining ones invariant. Specifically, we use Theorem 1 to solve the $\bm{Q}_{i}$ sub-problem. The obtained optimal solution results in the following inequality holds:
\begin{equation}\label{eq:30}
\begin{split}
&\mathcal{H}\Big(\{\bm{Q}_{i}^{(j+1)}, \bm{G}_{i}^{(j)}, \bm{\Lambda}_{i}^{(j)}, \bm{\Pi}_{i}^{(j)}, \phi_{i}^{(j)}\}_{i=1}^{v}, \bm{A}^{(j)}\Big)\\                                                                                                             
&\leq \mathcal{H}\Big(\{\bm{Q}_{i}^{(j)}, \bm{G}_{i}^{(j)}, \bm{\Lambda}_{i}^{(j)}, \bm{\Pi}_{i}^{(j)}, \phi_{i}^{(j)}\}_{i=1}^{v}, \bm{A}^{(j)}\Big),
\end{split}
\end{equation}
where $\bm{Q}_{i}^{(j+1)}$ denotes the $(j+1)$-th iteration of the variable $\bm{Q}_{i}$. A similar inequality holds for the variable  $\bm{A}$ as well, in accordance with Theorem 1.

The solution for $\bm{G}_{i}$ is determined by solving a projection capped simplex problem, which can be optimized to reach a global minimum~\cite{wang2015projection}. Consequently, a similar inequality to that in~(\ref{eq:30}) applies to the variable $\bm{G}_{i}$.

For the $\bm{\Pi}_{i}$ sub-problem, we determine its optimal solution through an exhaustive search method. This leads to a similar inequality to that in~(\ref{eq:30}) for the variable $\bm{\Pi}_{i}$. Additionally, the closed-form solutions for the variables $\bm{\Lambda}_{i}$ and $\phi_{i}$ also lead to inequalities similar to that in~(\ref{eq:30}).

Based on the above observations, we have:
\begin{align}\label{eq:31}
&\mathcal{H}\Big(\{\bm{Q}_{i}^{(j+1)},\bm{G}_{i}^{(j+1)},\bm{\Lambda}_{i}^{(j+1)},\bm{\Pi}_{i}^{(j+1)},\phi_{i}^{(j+1)}\}_{i=1}^{v},
\bm{A}^{(j+1)}\Big)\nonumber\\                                                                                                   
&\leq \mathcal{H}\Big(\{\bm{Q}_{i}^{(j)},\bm{G}_{i}^{(j)},\bm{\Lambda}_{i}^{(j)},\bm{\Pi}_{i}^{(j)},\phi_{i}^{(j)}\}_{i=1}^{v},
\bm{A}^{(j)}\Big),
\end{align}
which indicates that the objective function in Eq.~(\ref{eq:29}) is monotonically decreasing. Furthermore, since it is lower-bounded, we can conclude that Algorithm~\ref{alg:1} converges.
\section{Experiments}\label{sec4}
In this section, we validate the superiority of the proposed PAVuC-ATS against twelve baseline methods.
\subsection{Baseline Methods and Datasets}
\textbf{Baseline methods.}
We compare the proposed PAVuC-ATS with twelve baseline approaches, including LMVSC~\cite{kang2020large}, FPMVS-CAG~\cite{wang2021fast}, FMVACC~\cite{wang2022align}, FDAGF~\cite{zhang2023let}, RCAGL~\cite{10440580}, CAMVC~\cite{zhang2024learning}, AEVCMVC~\cite{liu2024learn}, PVC~\cite{huang2020partially}, MvCLN~\cite{yang2021partially}, CMVNMF~\cite{zhang2015constrained}, UPMGC-SM~\cite{wen2023unpaired}, and VuCG~\cite{cao2024view}. Among these methods, LMVSC, FPMVS-CAG, FMVACC, FDAGF, RCAGL, CAMVC, and AEVCMVC are bipartite graph-based approaches that employ various techniques to learn anchors and graphs. PVC and MvCLN address the PVuP by integrating representation learning and data alignment within a unified deep learning framework. In contrast, CMVNMF, UPMGC-SM, and VuCG tackle the VuP by restoring cross-view correspondences through different alignment mechanisms.

\noindent\textbf{Datasets.} We conduct experiments on six widely used datasets:~\textbf{Protein Fold Prediction}\footnote{\url{http://mkl.ucsd.edu/dataset/protein-fold-prediction/.}} (ProteinFold) consists of 12 views, with each view containing 694 protein domains categorized into 27 distinct clusters. \textbf{Wiki}\footnote{\url{http://www.svcl.ucsd.edu/projects/crossmodal/.}} consists of 10 semantic classes, encompassing a total of 2,866 image-text pairs. Images and text descriptions are treated as two separate views. \textbf{Caltech-101}\footnote{\url{https://data.caltech.edu/records/mzrjq-6wc02/.}} consists of 9,144 images collected from Google Images, covering 101 object categories as well as one background category. Six kinds of features are extracted from each image, including Gabor, WM, CENTRIST, HOG, GIST, and LBP, which are regarded as six different views. Caltech101-20 is a subset of the Caltech-101 dataset, comprising 20 categories with a total of 2,386 samples. \textbf{Reuters} \cite{glewis2004new} consists of 18,758 samples sourced from news articles of the Reuters news agency, covering six different categories. It includes the original English version as well as four translations (French, German, Spanish, and Italian), which are considered as five views. \textbf{CIFAR-10}\footnote{\url{https://www.cs.toronto.edu/kriz/cifar.html.}} consists of 50,000 small color images categorized into 10 different clusters. Table~\ref{tab:3} presents the statistics of the used datasets.

\begin{table}[h!]
\caption{The statistics of the used datasets.}
\label{tab:3}
\centering
\begin{tabular}{p{1.4cm}<{\centering}p{0.3cm}<{\centering}p{0.3cm}<{\centering}p{0.6cm}<{\centering}p{4.1cm}<{\centering}}
\hline
\multicolumn{1}{c}{\multirow{1}{*}{Dataset}} &\multirow{1}{*}{$v$} &\multirow{1}{*}{$k$}&\multirow{1}{*}{$n$} &\multirow{1}{*}{$d_{i}$}\\
\hline
\multirow{1}{*}{ProteinFold}      &12   &27   &694	       &27/27/27/27/27/27/27/27/27/27/27/27\\
\multirow{1}{*}{Caltech101-20}    &6    &20   &2,386	   &48/40/254/1,984/512/928\\
\multirow{1}{*}{Wiki}             &2    &10   &2,866	   &128/10\\
\multirow{1}{*}{Caltech-101}      &6    &102  &9,144       &48/40/254/1,984/512/928\\
\multirow{1}{*}{Reuters}          &5    &6    &18,758 	   &21,531/24,892/34,251/15,506/11,547\\
\multirow{1}{*}{CIFAR-10}          &3    &10   &50,000      &512/2,048/1,024\\
\hline
\end{tabular}
\end{table}
\subsection{Experimental Settings}
We implement the baseline methods by adhering to the recommended parameters and network structures specified by the original authors, and report the best results achieved in most cases. In brief, the parameter $\alpha$ for LMVSC is selected from the set $\{10^{-3},10^{-2},10^{-1},10^{0},10^{1}\}$, while the parameter $\lambda$ for FMVACC is chosen from $\{10^{-4},10^{0},10^{5}\}$. For FDAGF, the parameter $\alpha$ is selected from $\{10^{-5},10^{-1},10^{1},10^{3}\}$, and $\lambda$ from $\{10^{1},10^{3},10^{5}\}$. In RCAGL, the parameter $\lambda$ is chosen from $\{0,10^{0},10^{2},10^{3},10^{6}\}$. For CAMVC, the parameter $\alpha$ is selected from $\{10^{-3},10^{-2},10^{-1},10^{0},10^{1}\}$, and $\beta$ from $\{10^{-1},10^{0},10^{1},10^{2},10^{3}\}$. In AEVCMVC, the parameter $\gamma$ is chosen from $\{10^{-1},10^{0},10^{1},10^{2}\}$, and $\lambda$ from $\{10^{-4},10^{-2},10^{0},10^{2}\}$. For PVC, the parameter $\mu$ is selected from $\{10^{-2}, 10^{-1}, 10^{0}, 10^{1}, 10^{2}, 10^{3}\}$. In VuCG, the parameter $\lambda$ is chosen from $\{1,4,7,10\}$, and $\tau$ from $\{1.2,1.5,1.8,2\}$. Additionally, the parameter $\beta$ for CMVNMF is set to 1. In our PAVuC-ATS, there are three parameters, including the control parameter $\alpha$, the trade-off parameter $\mu$, and the number of anchors. We search the optimal value of $\alpha$ from $\{1.1,1.3,1.5,1.7,1.9,2\}$, and $\mu$ from $\{10^{-3},10^{-2},10^{-1},10^{0},10^{1}\}$. Moreover, the number of anchors is chosen from $\{1k,3k,5k\}$.

To create the fully unaligned multi-view datasets, we randomly shuffle the data within each view. We evaluate the baseline methods LMVSC, FPMVS-CAG, FMVACC, FDAGF, RCAGL, CAMVC, AEVCMVC, CMVNMF, UPMGC-SM, VuCG, as well as our PAVuC-ATS on these datasets. Furthermore, comparisons with PVC and MvCLN are conducted on the fully unaligned two-view datasets due to their two-view configuration, i.e., using views 9 and 12 for ProteinFold, views 2 and 5 for Caltech101-20 and Caltech-101, and views 1 and 2 for Wiki, Reuters and CIFAR-10 datasets. Since the two methods can only address the PVuP, we retain 1\% of the aligned data for them. To reduce computational demands, we project each data point from the Reuters dataset into a latent space with dimension 100 for all methods. For fairness, all algorithms are implemented on a PC with Intel(R) Core (TM) i7-8700 CPU @ 3.70GHz and 32.0GB RAM. Additionally, the average experimental results are reported based on the ten distinct shuffled versions of each dataset.

We evaluate the clustering performance using three widely used metrics in multi-view clustering tasks: accuracy (ACC), normalized mutual information (NMI), and F-score (F). Higher values indicate better clustering performance.
\begin{table*}[t!]
\caption{Performance comparison on the fully unaligned multi-view datasets: ProteinFold, Caltech101-20, Wiki, Caltech-101, Reuters, and CIFAR-10, where the best results are marked in bold, and '--' indicates out of memory.}
\label{tab:4}
\centering
\resizebox{7.2in}{!}{
\begin{tabular}{p{0.5cm}<{\centering}p{0.5cm}<{\centering}p{0.5cm}<{\centering}p{0.5cm}<{\centering}p{0.5cm}<{\centering}
p{0.5cm}<{\centering}p{0.5cm}<{\centering}p{0.5cm}<{\centering}p{0.5cm}<{\centering}p{0.5cm}<{\centering}p{0.5cm}<{\centering}
p{0.5cm}<{\centering}p{0.5cm}<{\centering}p{0.5cm}<{\centering}p{0.5cm}<{\centering}p{0.5cm}<{\centering}p{0.5cm}<{\centering}
p{0.5cm}<{\centering}p{0.5cm}<{\centering}p{0.5cm}<{\centering}}
\hline
\multirow{2}{*}{Method} &\multicolumn{3}{c}{ProteinFold} &\multicolumn{3}{c}{Caltech101-20} &\multicolumn{3}{c}{Wiki}&\multicolumn{3}{c}{Caltech-101}&\multicolumn{3}{c}{Reuters}&\multicolumn{3}{c}{CIFAR-10}\\\cline{2-4}\cline{5-7}
\cline{8-10}\cline{11-13}\cline{14-16}\cline{17-19}
&ACC &NMI &F &ACC &NMI &F &ACC &NMI &F &ACC &NMI &F &ACC &NMI &F &ACC &NMI &F\\
\hline
\multicolumn{1}{l}{LMVSC}     &14.09 &18.90 &6.64  &19.35 &10.27 &13.65 &12.92 &0.53  &10.83 &7.88  &17.13 &3.98 &29.51 &9.59  &25.95 &39.41  &33.33  &33.08\\
\multicolumn{1}{l}{FPMVS-CAG} &13.50 &16.12 &7.22  &12.78 &3.41  &11.74 &17.05 &3.45  &12.71 &6.09  &9.31  &3.93 &24.87 &2.01  &22.98 &23.43  &4.33   &13.82\\
\multicolumn{1}{l}{FMVACC}    &15.10 &20.80 &6.32  &14.54 &6.89  &10.12 &12.99 &0.83  &11.25 &7.63  &15.20 &4.26 &27.04 &3.60  &20.85 &41.57  &19.32  &22.96\\
\multicolumn{1}{l}{FDAGF}     &13.34 &15.19 &7.73  &14.71 &3.26  &14.03 &17.87 &5.06  &12.70 &6.55  &8.47  &4.32 &31.61 &9.18  &26.62 &59.54  &39.55  &42.57\\
\multicolumn{1}{l}{RCAGL}     &13.69 &10.91 &9.62  &29.30 &3.96  &25.86 &17.82 &4.39  &17.21 &9.45  &5.25  &5.47 &24.15 &0.68  &28.60 &25.92  &4.78   &13.01\\
\multicolumn{1}{l}{CAMVC}     &13.57 &19.24 &5.30  &9.19  &3.85  &8.14  &18.83 &5.23  &12.70 &6.59  &15.93 &2.66 &21.60 &1.54  &19.76 &47.35  &33.31  &26.97\\
\multicolumn{1}{l}{AEVCMVC}   &13.63 &16.40 &8.03  &22.20 &3.80  &20.35 &13.89 &1.50  &11.24 &5.12  &10.16 &3.15 &21.79 &0.07  &21.49 &24.61  &13.42  &21.61\\
\multicolumn{1}{l}{CMVNMF}    &13.33 &19.09 &5.85  &14.90 &11.68 &12.98 &33.68 &25.00 &27.03 &6.39  &17.42 &3.39 &27.67 &9.09  &25.61 &52.31  &44.06  &44.97\\
\multicolumn{1}{l}{UPMGC-SM}  &19.44 &26.13 &8.43  &26.48 &31.48 &19.27 &52.77 &50.96 &46.39 &11.10 &26.02 &6.60 &45.50 &30.15 &38.68
& --    & --    & --\\
\multicolumn{1}{l}{VuCG}      &21.85 &28.95 &11.02 &33.14 &30.00 &27.10 &51.30 &50.74 &43.55 &16.40 &32.66 &\textbf{12.72} &44.99    &31.63 &40.42 &85.64  &78.10  &77.12\\
\multicolumn{1}{l}{PAVuC-ATS} &\textbf{33.14} &\textbf{40.70} &\textbf{17.98} &\textbf{43.19} &\textbf{47.31} &\textbf{33.87} &\textbf{53.41} &\textbf{52.78} &\textbf{47.64} &\textbf{22.76} &\textbf{34.74} &7.41 &\textbf{55.74} &\textbf{39.90} &\textbf{45.48} &\textbf{90.17} &\textbf{79.57} &\textbf{81.86}\\
\hline
\end{tabular}}
\end{table*}
\begin{table*}[t!]
\caption{Performance comparison on the fully unaligned two-view datasets: ProteinFold, Caltech101-20, Wiki, Caltech-101, Reuters, and CIFAR-10, where the best results are marked in bold.}
\label{tab:5}
\centering
\resizebox{7.2in}{!}{
\begin{tabular}{p{0.5cm}<{\centering}p{0.5cm}<{\centering}p{0.5cm}<{\centering}p{0.5cm}<{\centering}p{0.5cm}<{\centering}
p{0.5cm}<{\centering}p{0.5cm}<{\centering}p{0.5cm}<{\centering}p{0.5cm}<{\centering}p{0.5cm}<{\centering}p{0.5cm}<{\centering}
p{0.5cm}<{\centering}p{0.5cm}<{\centering}p{0.5cm}<{\centering}p{0.5cm}<{\centering}p{0.5cm}<{\centering}p{0.5cm}<{\centering}
p{0.5cm}<{\centering}p{0.5cm}<{\centering}p{0.5cm}<{\centering}}
\hline
\multirow{2}{*}{Method} &\multicolumn{3}{c}{ProteinFold} &\multicolumn{3}{c}{Caltech101-20} &\multicolumn{3}{c}{Wiki}&\multicolumn{3}{c}{Caltech-101}&\multicolumn{3}{c}{Reuters}&\multicolumn{3}{c}{CIFAR-10}\\\cline{2-4}\cline{5-7}
\cline{8-10}\cline{11-13}\cline{14-16}\cline{17-19}
&ACC &NMI &F &ACC &NMI &F &ACC &NMI &F &ACC &NMI &F &ACC &NMI &F &ACC &NMI &F\\
\hline
\multicolumn{1}{l}{PVC}       &26.44 &35.88 &\textbf{26.57} &32.77 &36.52 &\textbf{37.64} &14.17 &2.64  &11.94 &12.33 &30.46 &\textbf{14.65} &39.84 &12.66 &39.09 &37.77 &12.04 &37.55\\
\multicolumn{1}{l}{MvCLN}     &14.12 &16.95 &10.49 &27.37 &31.20 &21.30 &14.45 &1.67  &13.72 &8.38  &23.27 &6.69  &37.24 &16.24 &31.34 &78.53 &70.96 &76.22\\
\multicolumn{1}{l}{PAVuC-ATS} &\textbf{32.96} &\textbf{41.37} &18.83 &\textbf{38.95} &\textbf{44.05} &33.02 &\textbf{53.41} &\textbf{52.78} &\textbf{47.64} &\textbf{21.91} &\textbf{34.57} &7.26  &\textbf{47.99} &\textbf{31.48} &\textbf{41.01} &\textbf{90.43} &\textbf{79.98} &\textbf{82.29}\\
\hline
\end{tabular}}
\end{table*}
\begin{figure*}[h!]
\begin{center}
\includegraphics[width=0.23\linewidth]{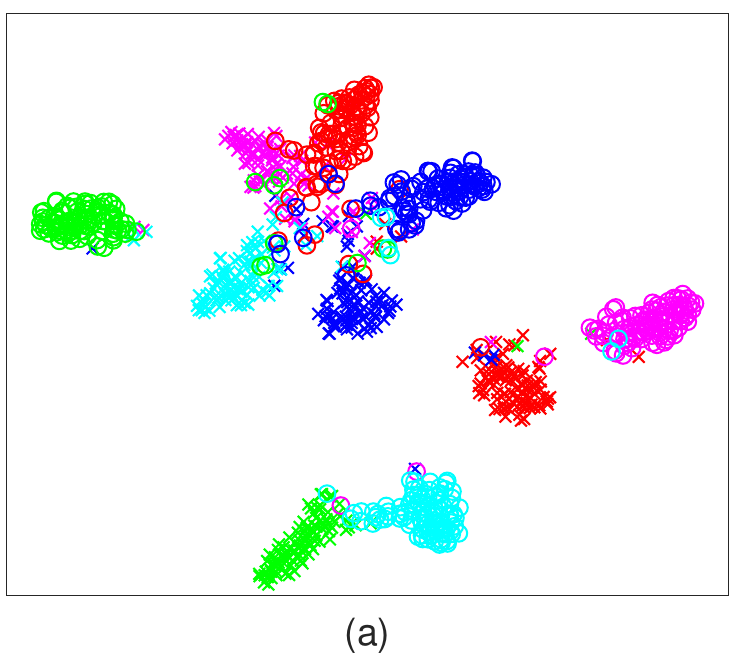}~~
\includegraphics[width=0.23\linewidth]{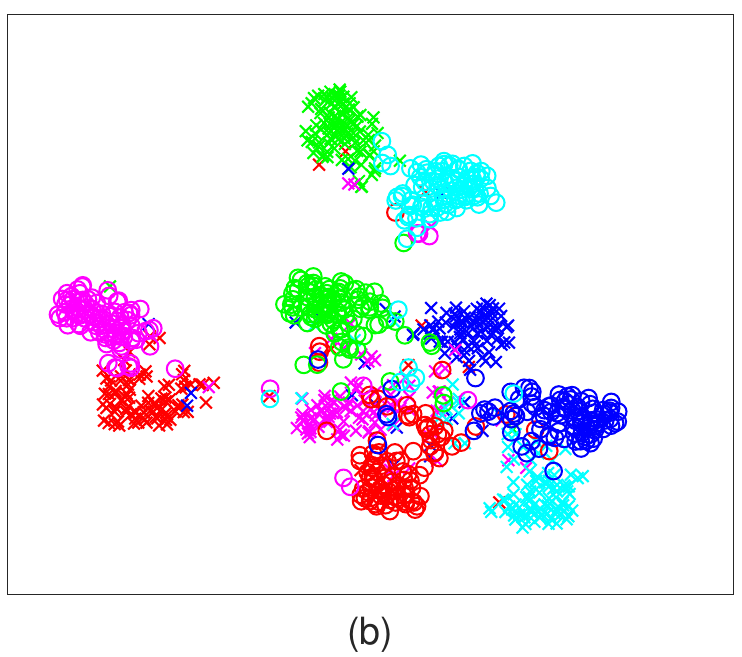}~~
\includegraphics[width=0.23\linewidth]{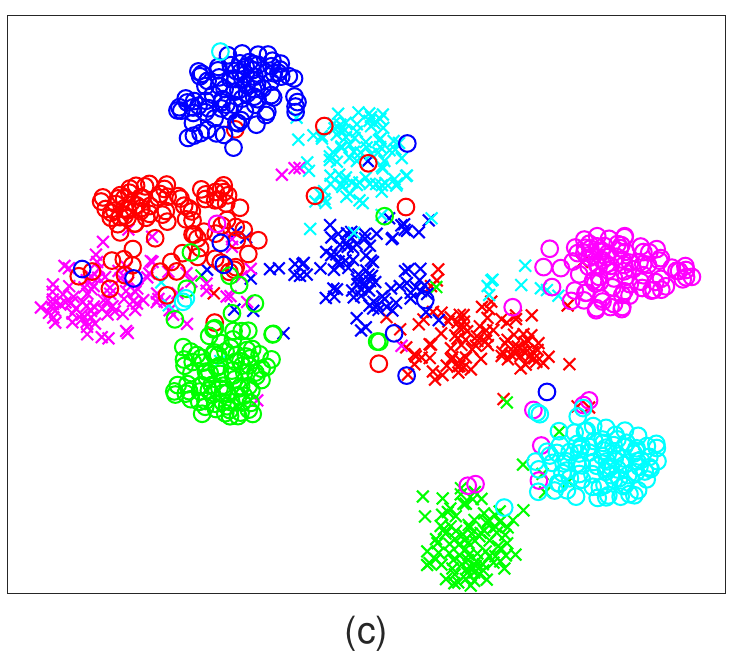}~~
\includegraphics[width=0.27\linewidth]{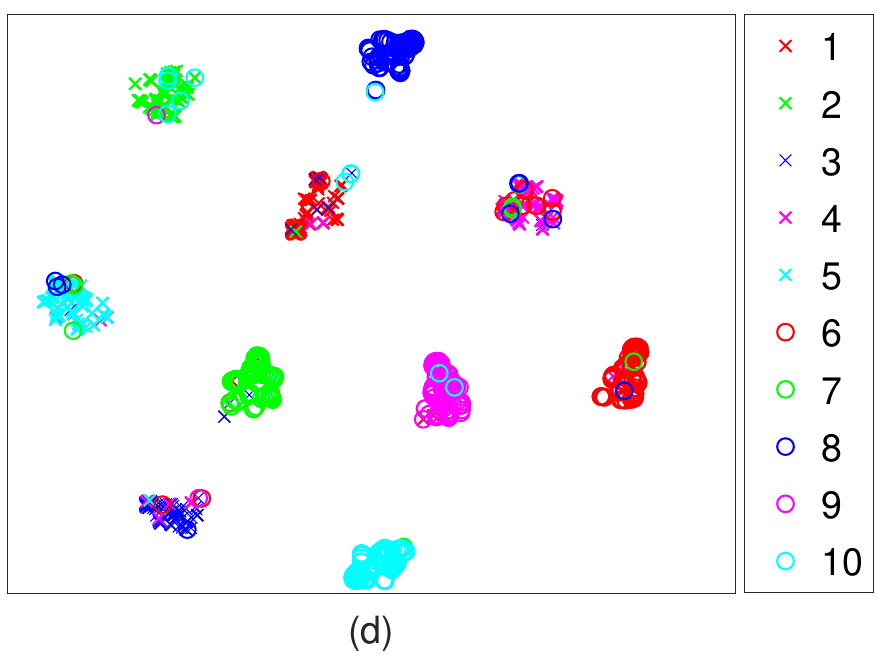}\\
\end{center}
\caption{Visualizations of (a) $\bm{X}_{1}$, (b) $\bm{X}_{2}$, (c) $\bm{X}_{3}$, and (d) $\bar{\bm{G}}$ on a subset of the CIFAR-10 dataset with an alignment ratio of $\rho=0$.}
\label{fig:3}
\end{figure*}
\begin{table*}[t!]
\caption{Performance comparison on the fully unaligned multi-view datasets: ProteinFold, Caltech101-20, Wiki, Caltech-101, Reuters, and CIFAR-10.}
\label{tab:6}
\centering
\resizebox{7.2in}{!}{
\begin{tabular}{p{0.5cm}<{\centering}p{0.5cm}<{\centering}p{0.5cm}<{\centering}p{0.5cm}<{\centering}p{0.5cm}<{\centering}
p{0.5cm}<{\centering}p{0.5cm}<{\centering}p{0.5cm}<{\centering}p{0.5cm}<{\centering}p{0.5cm}<{\centering}p{0.5cm}<{\centering}
p{0.5cm}<{\centering}p{0.5cm}<{\centering}p{0.5cm}<{\centering}p{0.5cm}<{\centering}p{0.5cm}<{\centering}p{0.5cm}<{\centering}
p{0.5cm}<{\centering}p{0.5cm}<{\centering}p{0.5cm}<{\centering}}
\hline
\multirow{2}{*}{Method} &\multicolumn{3}{c}{ProteinFold} &\multicolumn{3}{c}{Caltech101-20} &\multicolumn{3}{c}{Wiki}&\multicolumn{3}{c}{Caltech-101}&\multicolumn{3}{c}{Reuters}&\multicolumn{3}{c}{CIFAR-10}\\\cline{2-4}\cline{5-7}
\cline{8-10}\cline{11-13}\cline{14-16}\cline{17-19}
&ACC &NMI &F &ACC &NMI &F &ACC &NMI &F &ACC &NMI &F &ACC &NMI &F &ACC &NMI &F\\
\hline
\multicolumn{1}{l}{PAVuC-ATS} &33.14 &40.70 &17.98 &43.19 &47.31 &33.87 &53.41 &52.78 &47.64 &22.76 &34.74 &7.41 &55.74 &39.90 &45.48 &90.17 &79.57 &81.86\\
\multicolumn{1}{l}{NPA-VuC}   &16.58 &23.84 &7.06  &20.19 &3.58  &19.33 &12.90 &0.75  &10.61 &6.97  &10.78 &4.99 &29.12 &4.91  &22.36
 &39.37 &17.56 &21.38\\
\hline
\multicolumn{1}{l}{Difference}&16.56 &16.86 &10.92 &23.00 &43.73 &14.54 &40.51 &52.03 &37.03 &15.79 &23.96 &2.42 &26.62 &34.99 &23.12   &50.80 &62.01 &60.48
\\
\hline
\end{tabular}}
\end{table*}
\subsection{Experimental Results}
\label{sec5}
We validate the superiority of the proposed PAVuC-ATS against twelve baseline methods on six fully unaligned real datasets. Among these methods, LMVSC, FPMVS-CAG, FMVACC, FDAGF, RCAGL, CAMVC, and AEVCMVC focus on the CA-MVC, while PVC, MvCLN, CMVNMF, UPMGC-SM, VuCG, as well as our PAVuC-ATS address the VuP. The experimental results are presented in Tables~\ref{tab:4} and~\ref{tab:5}, from which we can draw the following observations.
\begin{itemize}
\item In the fully unaligned multi-view scenarios, Table~\ref{tab:4} indicates that the proposed PAVuC-ATS consistently outperforms the compared methods. Specifically, the evaluation metric ACC of PAVuC-ATS exceeds that of the second best method by 11.29\%, 10.05\%, 0.64\%, 6.36\%, 10.24\%, and 4.53\% on the ProteinFold, Caltech101-20, Wiki, Caltech-101, Reuters and CIFAR-10 datasets, respectively. Similarly, the NMI metric shows improvements of 11.75\%, 15.83\%, 1.82\%, 2.08\%, 8.27\%, and 1.47\% for the same datasets.
\item Among the compared methods, the CA-MVC approaches exhibit subpar performance on all six fully unaligned datasets. This is primarily because the derivation of a consistent similarity matrix across multiple views relies on cross-view correspondences, which are absent in fully unaligned datasets, ultimately degrading their performance. This observation underscores the significance of the alignment mechanism in dealing with view-unaligned data.
\item In the fully unaligned two-view scenarios, as presented in Table~\ref{tab:5}, the evaluation metric ACC of PAVuC-ATS surpasses that of the second best method by 6.52\%, 6.18\%, 38.96\%, 9.58\%, 8.15\%, and 11.90\%, and the NMI by 5.49\%, 7.53\%, 50.14\%, 4.11\%, 15.24\%, and 9.02\%, on the ProteinFold, Caltech101-20, Wiki, Caltech-101, Reuters and CIFAR-10 datasets, respectively.
\item We visualize the feature matrices $\{\bm{X}_{i}\}_{i=1}^{v}$ and the fused anchor graph $\bar{\bm{G}}$ defined in Eq.~(\ref{eq:28}), utilizing the t-SNE algorithm on a subset of the CIFAR-10 dataset with an alignment ratio of $\rho=0$, where the subset comprises 100 randomly selected samples from each category of the CIFAR-10 dataset, totaling 1,000 samples. Compared to the scatter plots representing the feature maxtrices $\{\bm{X}_{i}\}_{i=1}^{v}$, the visualization of the fused anchor graph $\bar{\bm{G}}$ depicted in Fig.~\ref{fig:3}(d) indicates that the proposed method can effectively segment the fully view-unaligned data.
\end{itemize}
\subsection{Ablation Study}
In this section, we provide an effectiveness validation for the proposed probabilistic alignment mechanism on six fully unaligned datasets. To this end, we introduce a variant of PAVuC-ATS, denoted by NPA-VuC, where the alignment term in Eq.~(\ref{eq:7}) is removed. The significant performance difference between PAVuC-ATS and NPA-VuC, as tabulated in Table~\ref{tab:6}, confirms the powerful capability of the proposed probabilistic alignment mechanism in handling view-unaligned data.
\begin{figure*}[t!]
\begin{center}
\includegraphics[width=0.31\linewidth]{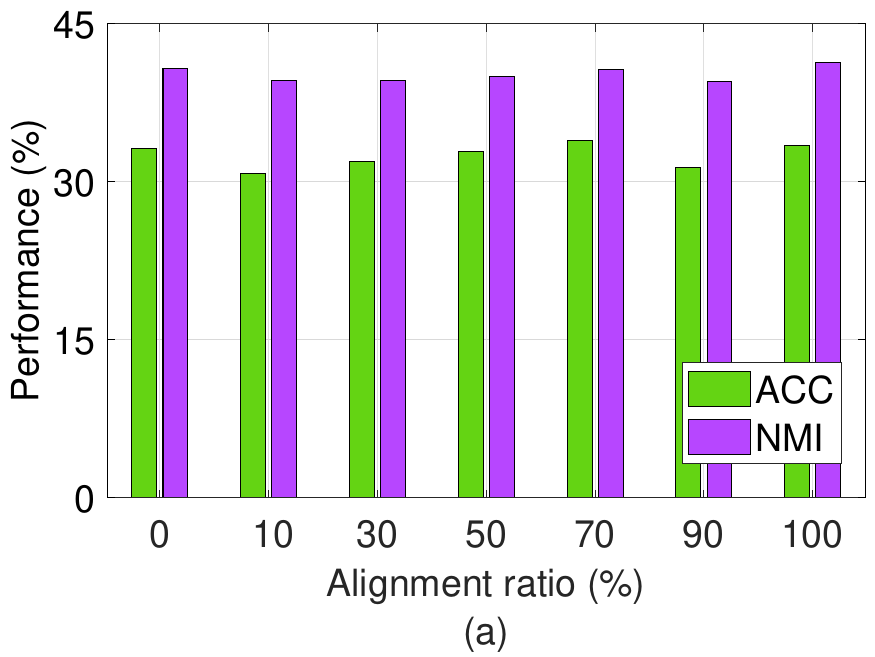}~~
\includegraphics[width=0.31\linewidth]{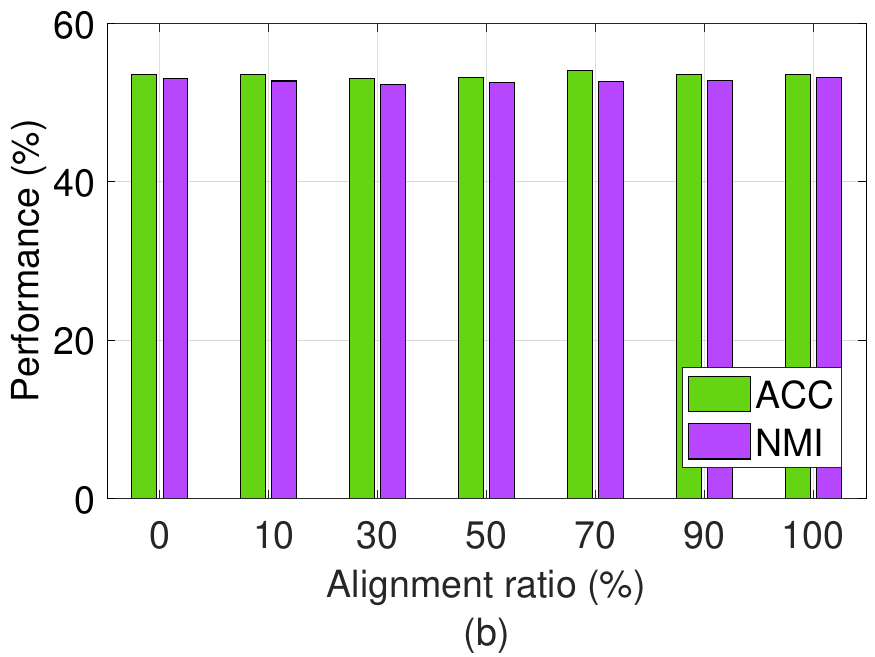}~~
\includegraphics[width=0.31\linewidth]{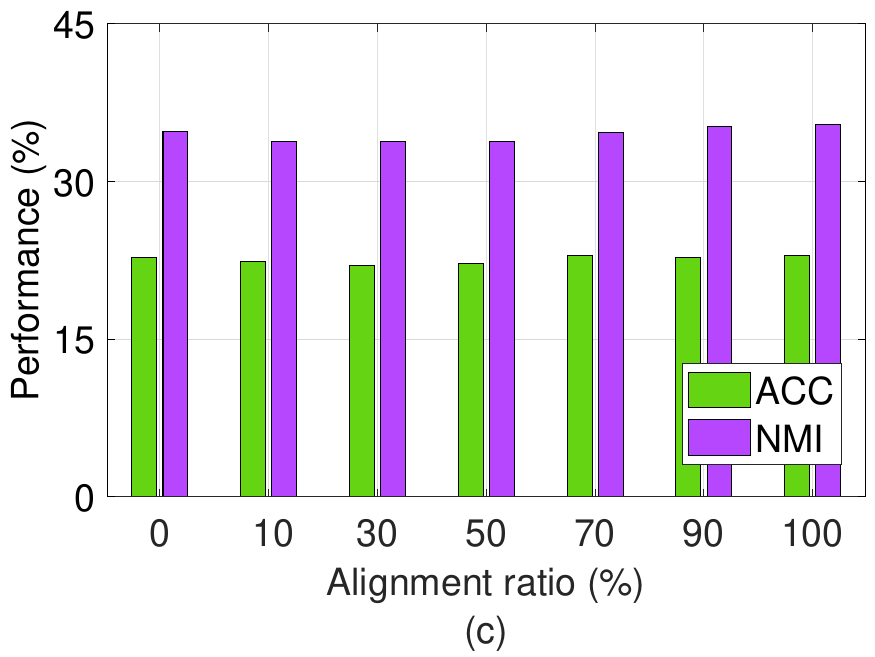}
\end{center}
\caption{The evaluation metrics (ACC and NMI) versus the alignment ratio on the fully unaligned multi-view datasets: (a) ProteinFold, (b) Wiki, and (c) Caltech-101.}
\label{fig:4}
\end{figure*}
\begin{figure*}[h!]
\begin{center}
\includegraphics[width=0.31\linewidth]{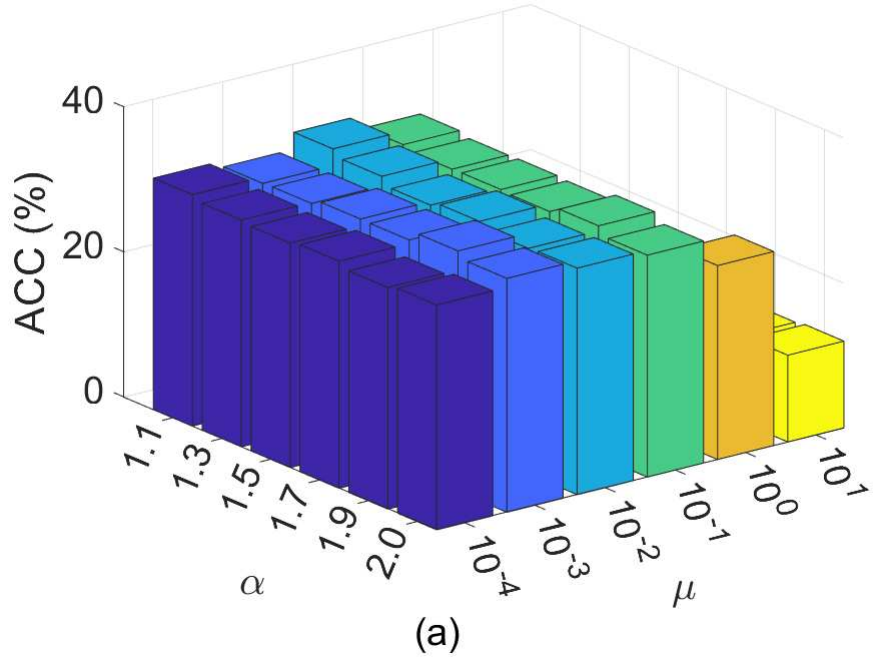}~~
\includegraphics[width=0.31\linewidth]{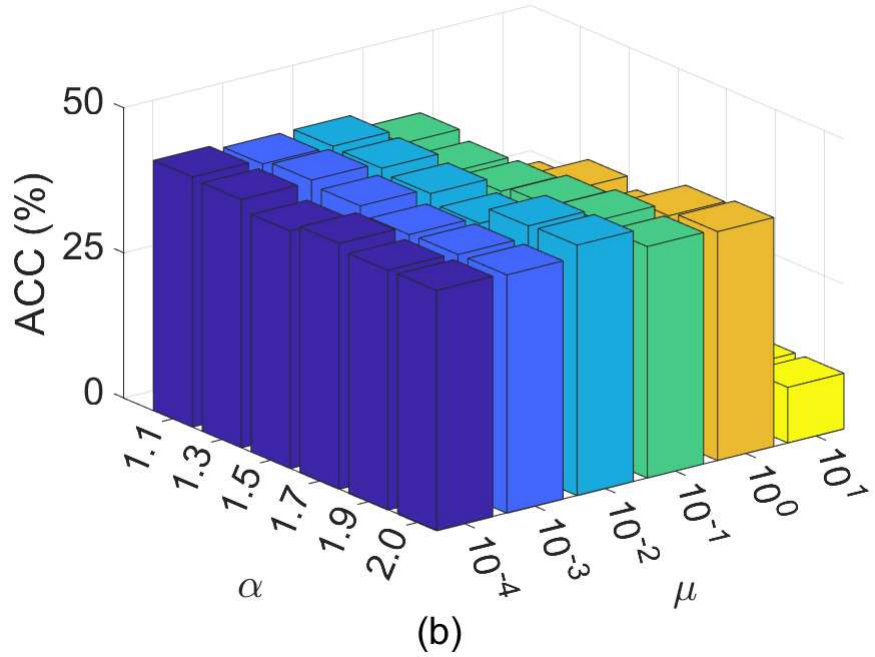}~~
\includegraphics[width=0.31\linewidth]{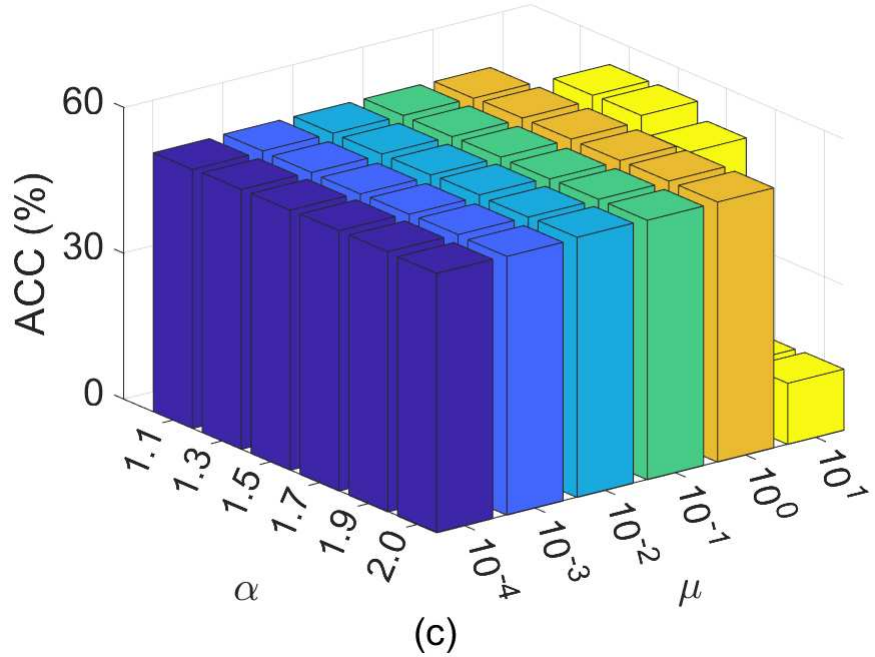}\\
\includegraphics[width=0.31\linewidth]{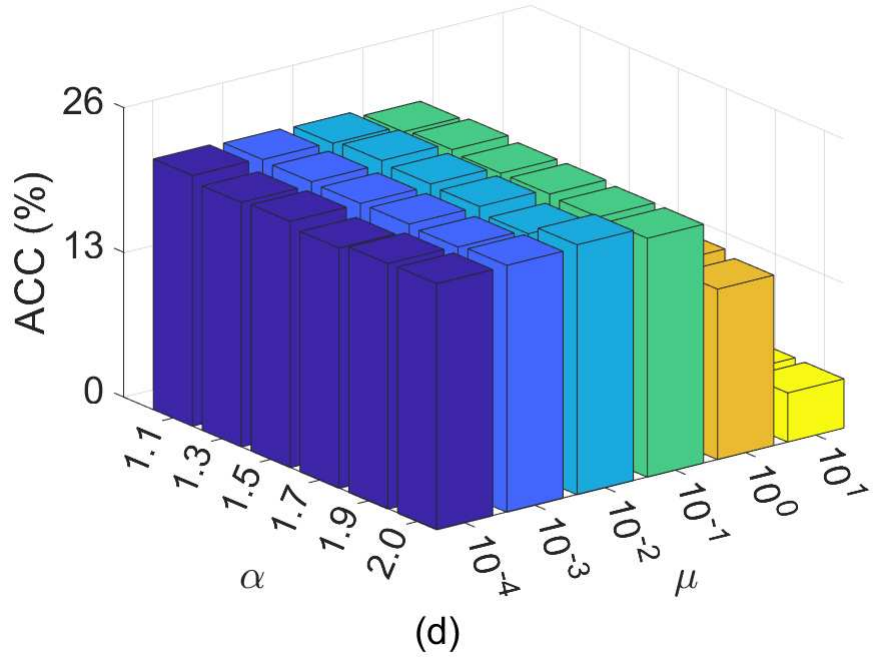}~~
\includegraphics[width=0.31\linewidth]{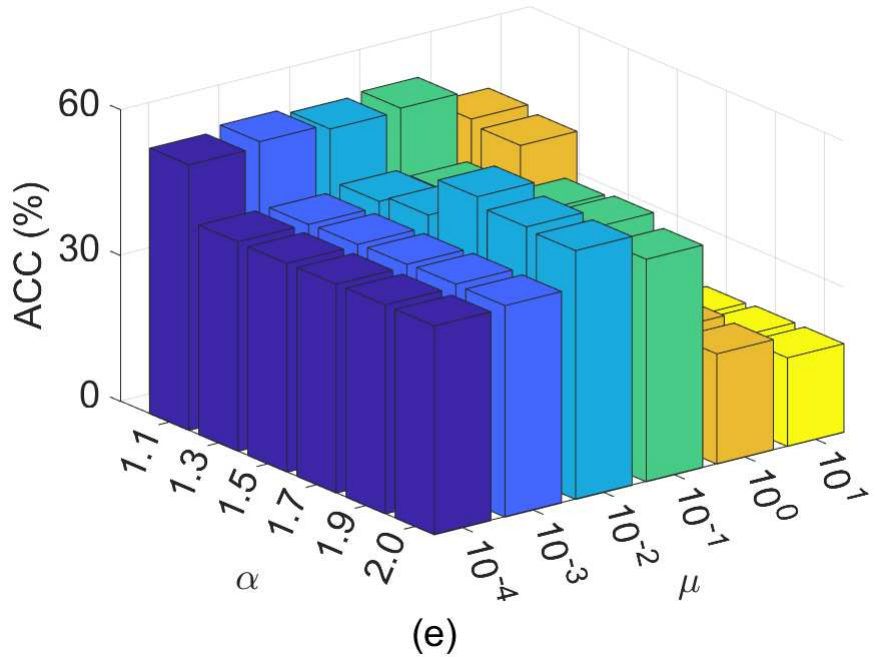}~~
\includegraphics[width=0.31\linewidth]{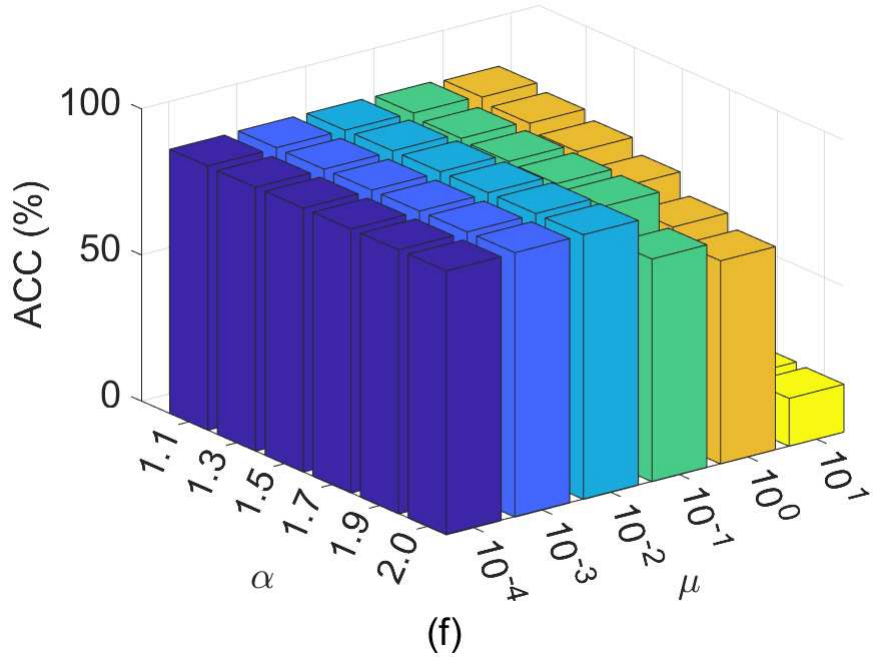}\\
\includegraphics[width=0.31\linewidth]{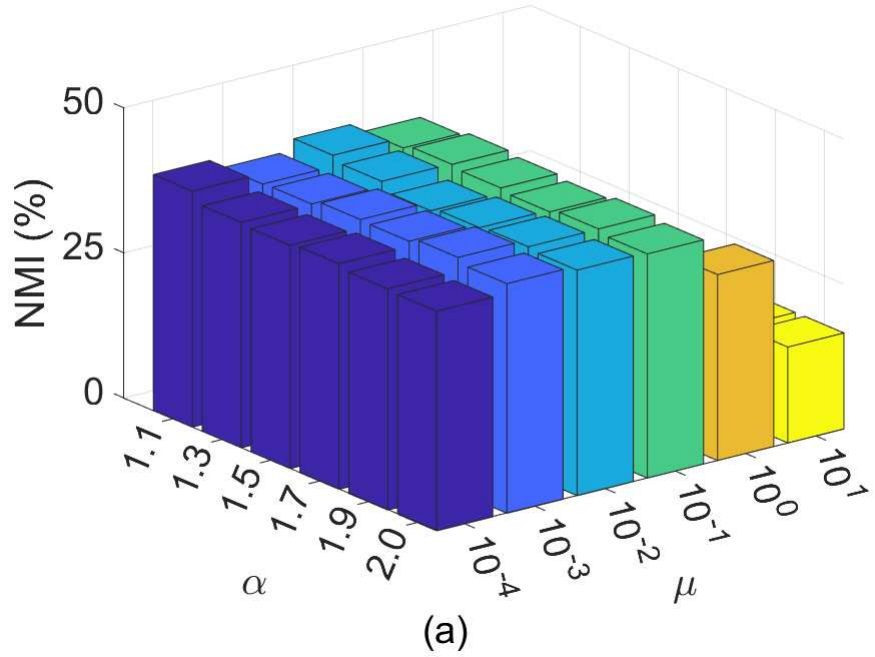}~~
\includegraphics[width=0.31\linewidth]{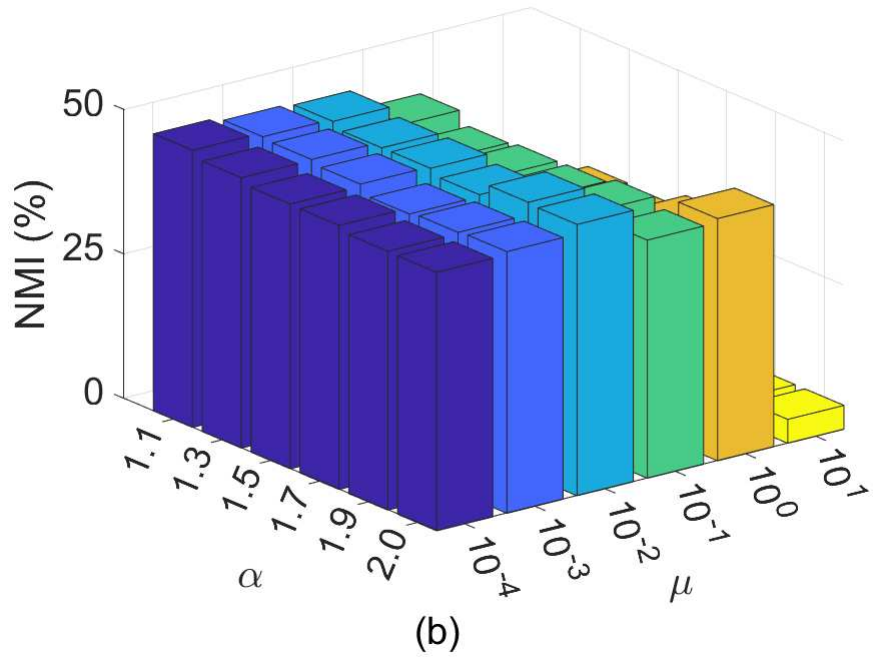}~~
\includegraphics[width=0.31\linewidth]{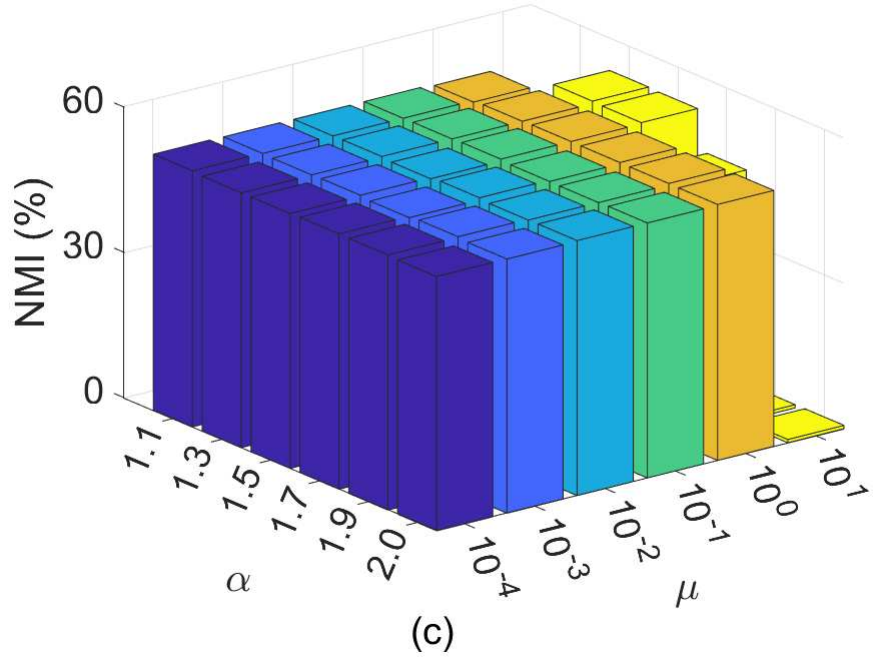}\\
\includegraphics[width=0.31\linewidth]{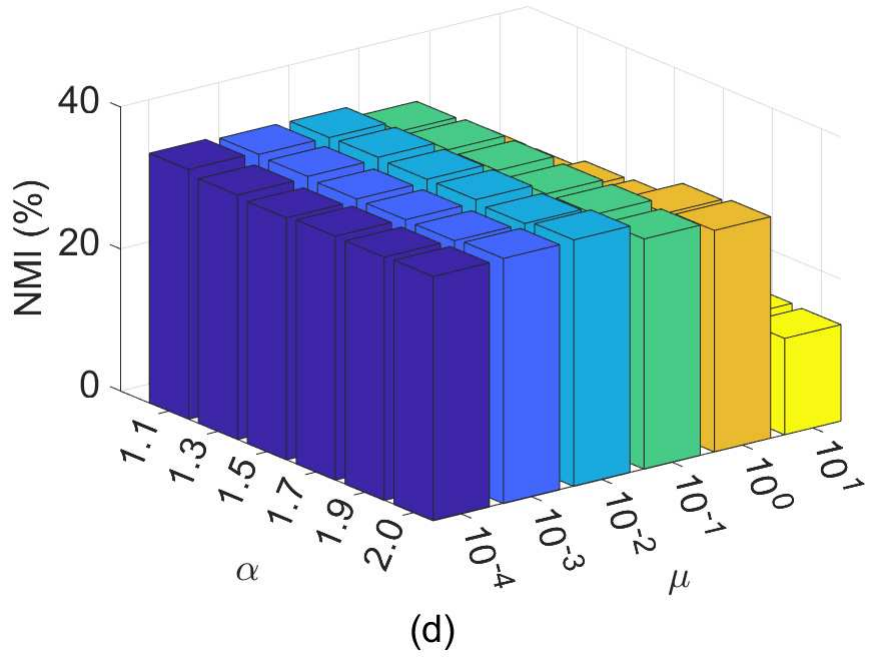}~~
\includegraphics[width=0.31\linewidth]{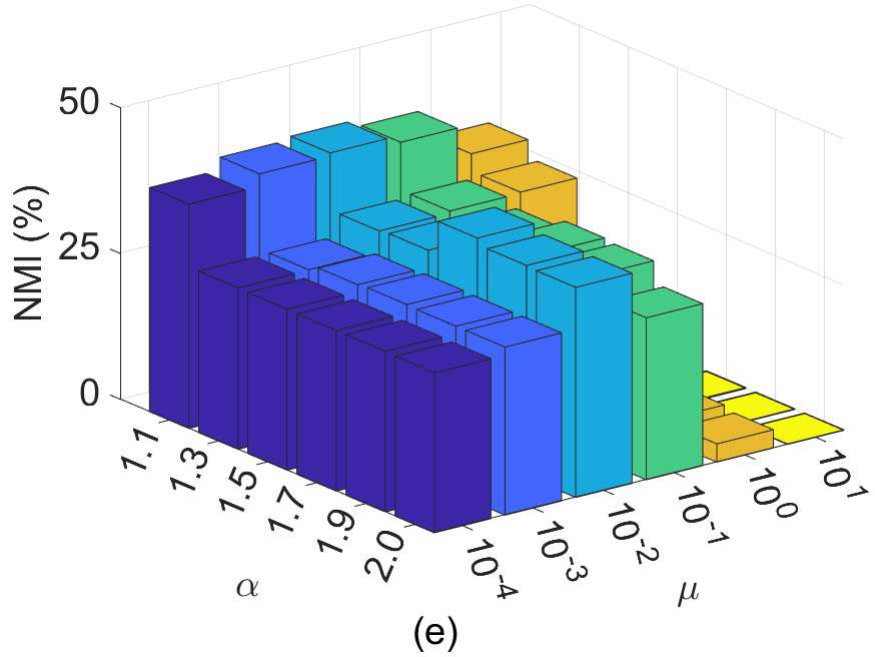}~~
\includegraphics[width=0.31\linewidth]{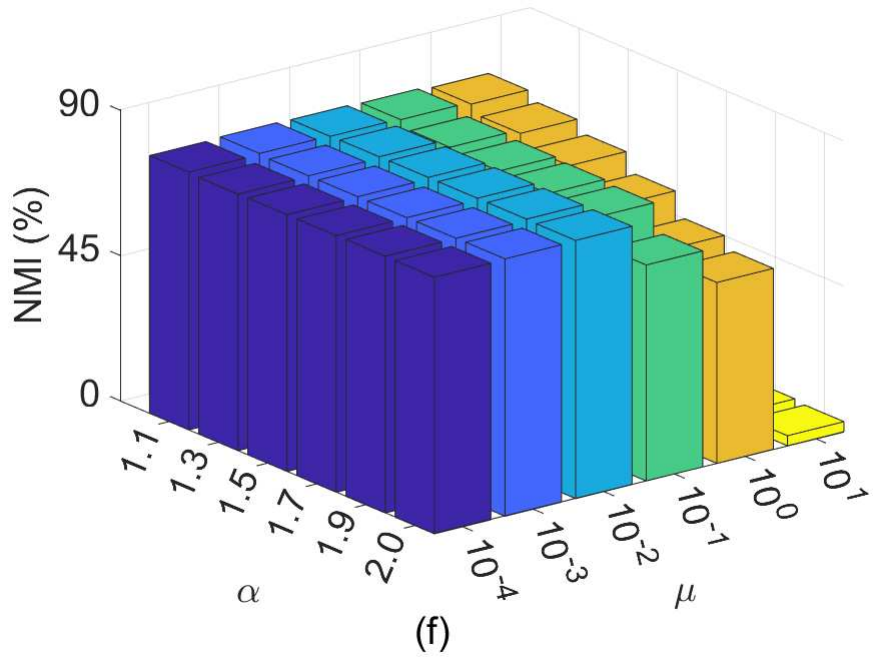}\\
\end{center}
\caption{The evaluation metrics (ACC and NMI) versus the parameters $\alpha$ and $\mu$ on the fully unaligned multi-view datasets: (a) ProteinFold, (b) Caltech101-20, (c) Wiki, (d) Caltech-101, (e) Reuters, and (f) CIFAR-10.}
\label{fig:5}
\end{figure*}
\begin{figure*}[t!]
\begin{center}
\includegraphics[width=0.31\linewidth]{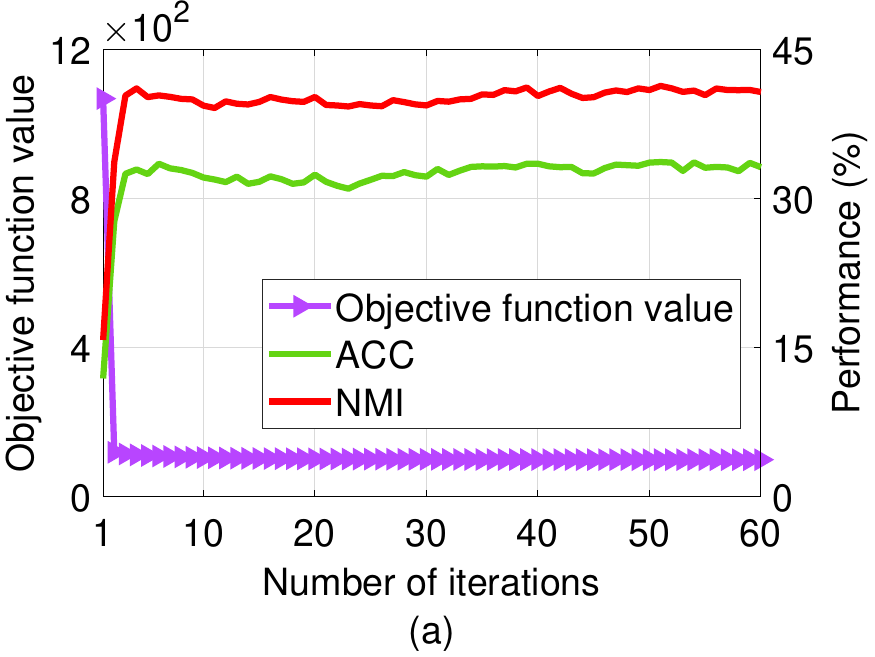}~~
\includegraphics[width=0.31\linewidth]{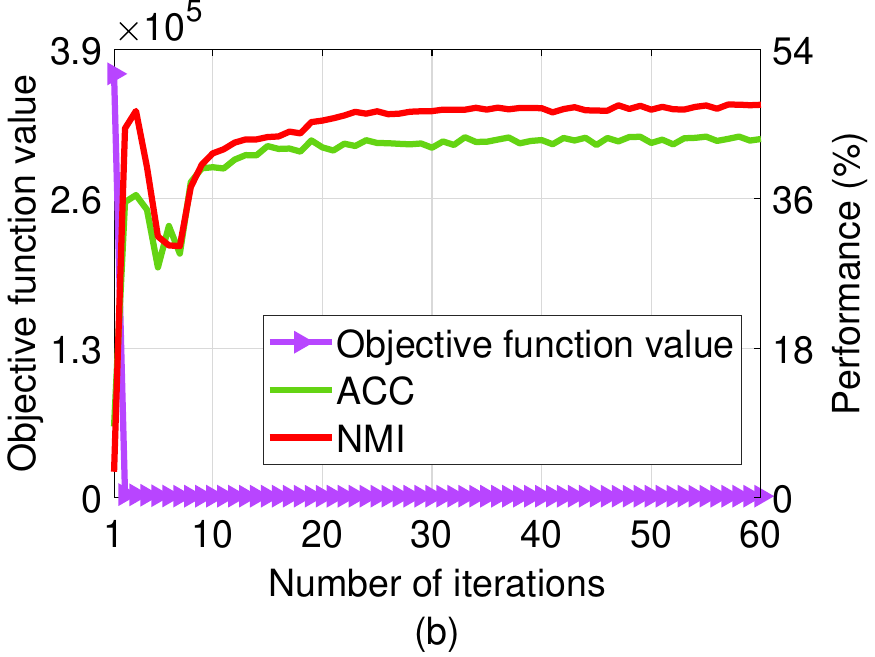}~~
\includegraphics[width=0.31\linewidth]{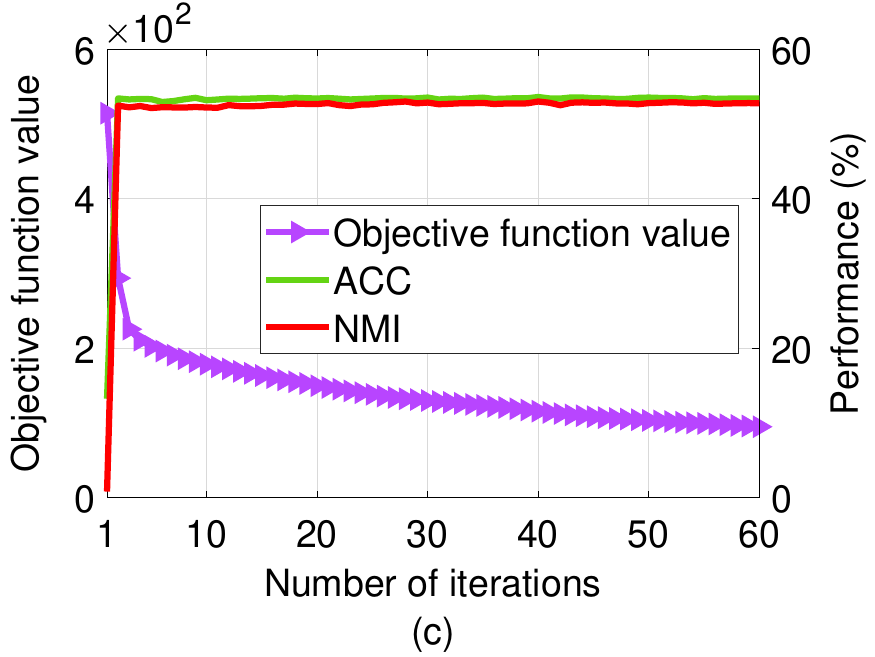}\\
\includegraphics[width=0.31\linewidth]{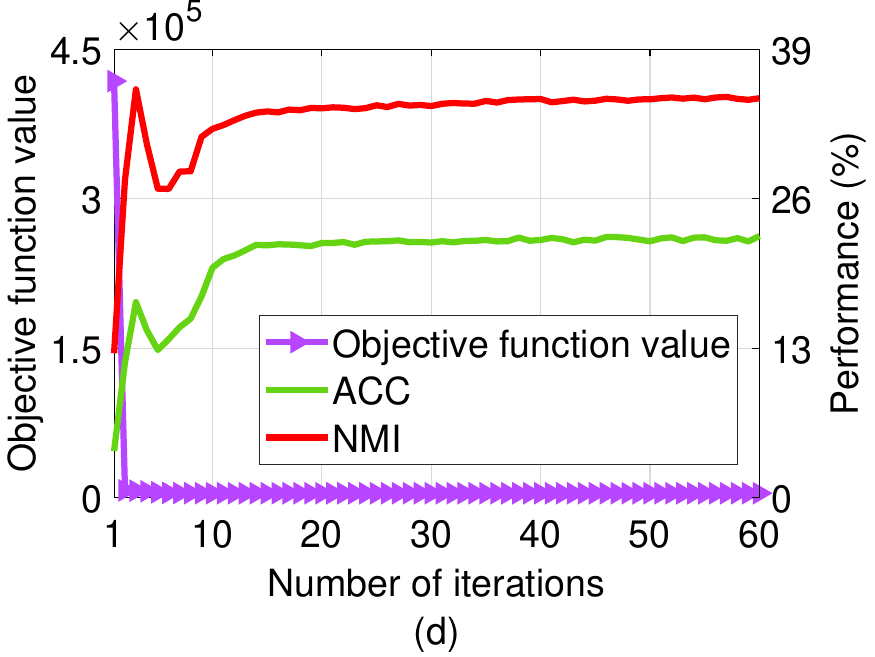}~~
\includegraphics[width=0.31\linewidth]{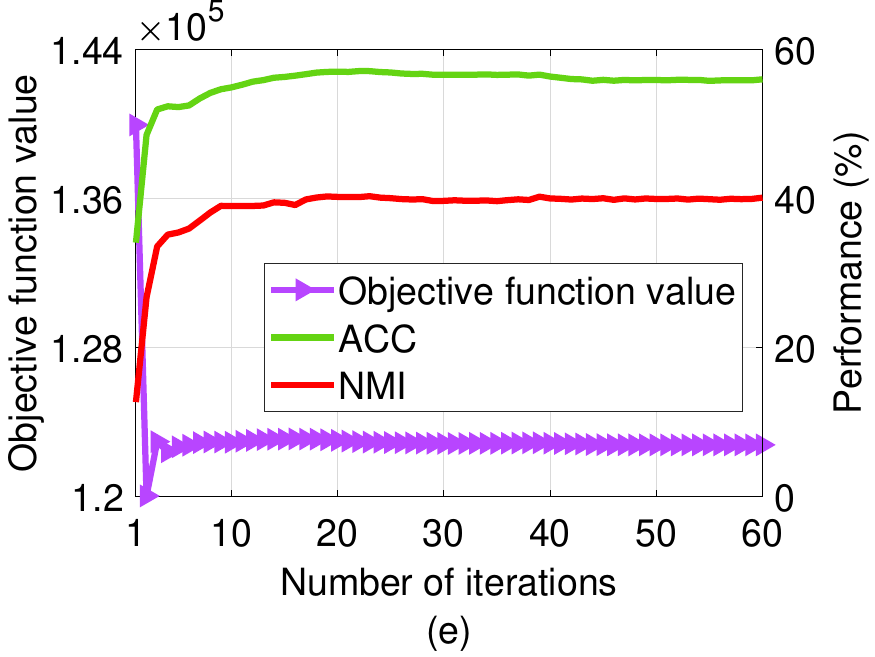}~~
\includegraphics[width=0.31\linewidth]{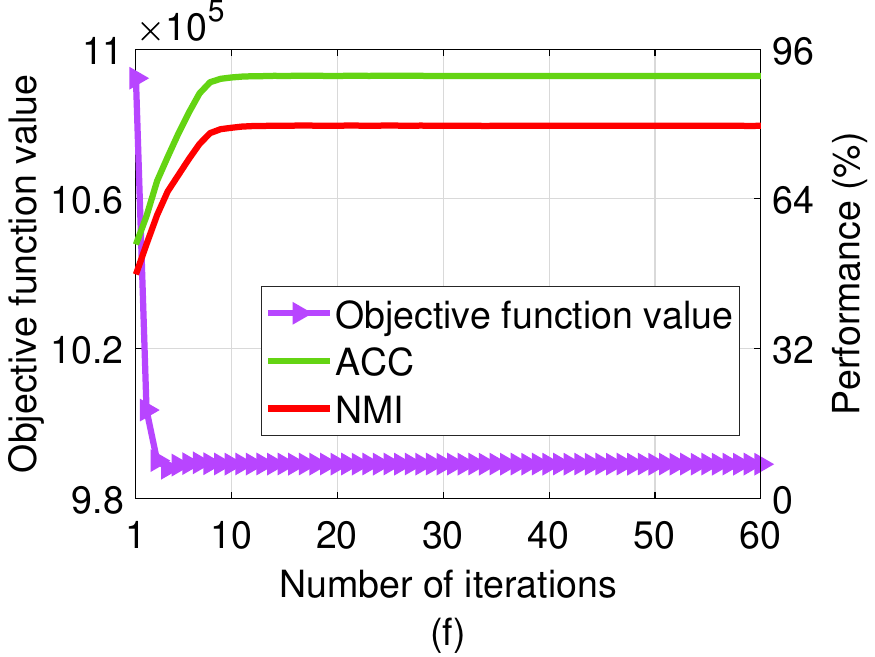}
\end{center}
\caption{The objective function values and evaluation metrics (ACC and NMI) versus the number of iterations on the fully unaligned multi-view datasets: (a) ProteinFold, (b) Caltech101-20, (c) Wiki, (d) Caltech-101, (e) Reuters, and (f) CIFAR-10.}
\label{fig:6}
\end{figure*}
\subsection{Effect of Alignment Ratios}
The goal of this experiment is to evaluate the effect of alignment ratios on clustering performance. We conduct experiments on the fully unaligned ProteinFold, Wiki, and Caltech-101 datasets, and explore the alignment ratio $\rho$ within the range $\{0, 10\%, 30\%, 50\%, 70\%, 90\%, 100\%\}$. Fig.~\ref{fig:4} reveals intriguing insights, where the evaluation metrics (ACC and NMI) do not strictly escalate with the augmentation of the alignment ratio, but rather exhibit certain fluctuations. This anomaly stems from the fact that the composition of aligned and unaligned samples shifts dynamically with the adjustment of the alignment ratio. Moreover, the evaluation metrics (ACC and NMI) reach their peak values at alignment ratios of either 70\% or 100\%.
\subsection{Sensitivity Analysis}
In this section, we conduct a sensitivity analysis for the control parameter $\alpha>1$ and the trade-off parameter $\mu>0$ on six fully unaligned datasets. The evaluation metrics ACC and NMI are respectively regarded as the functions of $\alpha$ and $\mu$. As depicted in Fig.~\ref{fig:5}, the evaluation metrics ACC and NMI remain rather stable on all six datasets when the parameters $\alpha$ and $\mu$ are varied within the specified ranges of $\{1.1, 1.3, 1.5, 1.7, 1.9, 2\}$ and $\{10^{-4}, 10^{-3}, 10^{-2}, 10^{-1}\}$, respectively.
\subsection{Time Comparison}
We compare the computational time of the proposed PAVuC-ATS with the baseline methods CMVNMF, UPMGC-SM, and VuCG on six fully unaligned datasets that address the VuP. We exclude the partially view-aligned clustering methods PVC and MvCLN from this time comparison, since they are based on deep learning. Table~\ref{tab:7} indicates that the baseline CMVNMF is very fast due to its application of the non-negative matrix factorization (NMF) method. In contrast to the other baseline methods, our PAVuC-ATS is highly efficient on all six datasets.
\begin{table}[h!]
\caption{Computational time (s) comparison, where '--' indicates out of memory.}
\label{tab:7}
\centering
\begin{tabular}{p{1.7cm}<{\centering}p{1.2cm}<{\centering}p{1.2cm}<{\centering}p{1.2cm}<{\centering}p{1.4cm}<{\centering}}
\hline
\multicolumn{1}{c}{\multirow{1}{*}{Dataset}} &\multirow{1}{*}{CMVNMF} &\multirow{1}{*}{UPMGC-SM}&\multirow{1}{*}{VuCG} &\multirow{1}{*}{PAVuC-ATS}\\
\hline
\multirow{1}{*}{ProteinFold}      &1.4       &32.6      &6.9	   &2.6\\
\multirow{1}{*}{Caltech101-20}    &7.5       &280.6     &38.9	   &29.8\\
\multirow{1}{*}{Wiki}             &16.3      &87.8      &22.3	   &7.4\\
\multirow{1}{*}{Caltech-101}      &134.9     &4,767.9   &809.8     &342.9\\
\multirow{1}{*}{Reuters}          &374.3     &9,286.0   &4,942.9   &639.4\\
\multirow{1}{*}{CIFAR-10}         &3,055.9   &--        &44,260.0  &4,157.6\\
\hline
\end{tabular}
\end{table}
\subsection{Convergence Validation}
In this experiment, we investigate the convergence behavior of Algorithm~\ref{alg:1} on six fully unaligned datasets.
As illustrated in Fig.~\ref{fig:6}, within 60 iterations, the curves of the evaluation metrics ACC and NMI undergo an initial swift rise, subsequently tending towards stabilization. Meanwhile, the trajectory of the objective function values experiences an initial steep decline, ultimately reaching a local minimum. This empirical observation provides strong evidence that supports the theoretical analysis of Algorithm~\ref{alg:1} presented in Section~\ref{sec:3.4}.
\section{Conclusions}
In this paper, we propose an efficient and effective clustering solution for the VuP with arbitrary alignment levels by incorporating a permutation derivation procedure into the bipartite graph framework, in which we learn cross-view anchors and view-specific graphs employing the bipartite graph, and derive the permutations applied to the unaligned graphs through a probabilistic alignment mechanism. The integration of anchor graph learning and the probabilistic alignment mechanism enhances the performance while maintaining high scalability. Extensive experiments conducted on six real datasets validate the effectiveness of the proposed model and methodology. In the future, we aim to further explore a more general framework for the diversified view issues and potential applications of the proposed method.
\ifCLASSOPTIONcompsoc
  \section*{Acknowledgments}
This work is supported by the National Natural Science Foundation of China (62020106012, U1836218, 61672265), the 111 Project of Ministry of Education of China (B12018), and the Engineering and Physical Sciences Research Council (EPSRC) (EP/N007743/1, MURI/EPSRC/DSTL, EP/R018456/1).
\else
  \section*{Acknowledgment}

\fi
\ifCLASSOPTIONcaptionsoff
  \newpage
\fi

\bibliographystyle{IEEEtran}
\bibliography{mybib}

\begin{thebibliography}{10}
\providecommand{\url}[1]{#1}
\csname url@samestyle\endcsname
\providecommand{\newblock}{\relax}
\providecommand{\bibinfo}[2]{#2}
\providecommand{\BIBentrySTDinterwordspacing}{\spaceskip=0pt\relax}
\providecommand{\BIBentryALTinterwordstretchfactor}{4}
\providecommand{\BIBentryALTinterwordspacing}{\spaceskip=\fontdimen2\font plus
\BIBentryALTinterwordstretchfactor\fontdimen3\font minus
  \fontdimen4\font\relax}
\providecommand{\BIBforeignlanguage}[2]{{%
\expandafter\ifx\csname l@#1\endcsname\relax
\typeout{** WARNING: IEEEtran.bst: No hyphenation pattern has been}%
\typeout{** loaded for the language `#1'. Using the pattern for}%
\typeout{** the default language instead.}%
\else
\language=\csname l@#1\endcsname
\fi
#2}}
\providecommand{\BIBdecl}{\relax}
\BIBdecl

\bibitem{bickel2004multi}
S.~Bickel and T.~Scheffer, ``Multi-view clustering,'' in \emph{ICDM}, vol.~4,
  no. 2004.\hskip 1em plus 0.5em minus 0.4em\relax Citeseer, 2004, pp. 19--26.

\bibitem{zhang2018generalized}
C.~Zhang, H.~Fu, Q.~Hu, X.~Cao, Y.~Xie, D.~Tao, and D.~Xu, ``Generalized latent
  multi-view subspace clustering,'' \emph{IEEE Transactions on Pattern Analysis
  and Machine Intelligence}, vol.~42, no.~1, pp. 86--99, 2018.

\bibitem{li2019reciprocal}
R.~Li, C.~Zhang, H.~Fu, X.~Peng, T.~Zhou, and Q.~Hu, ``Reciprocal multi-layer
  subspace learning for multi-view clustering,'' in \emph{Proceedings of the
  IEEE/CVF International Conference on Computer Vision}, 2019, pp. 8172--8180.

\bibitem{li2020multiview}
X.~Li, H.~Zhang, R.~Wang, and F.~Nie, ``Multiview clustering: A scalable and
  parameter-free bipartite graph fusion method,'' \emph{IEEE Transactions on
  Pattern Analysis and Machine Intelligence}, vol.~44, no.~1, pp. 330--344,
  2020.

\bibitem{xia2022tensorized}
W.~Xia, Q.~Gao, Q.~Wang, X.~Gao, C.~Ding, and D.~Tao, ``Tensorized bipartite
  graph learning for multi-view clustering,'' \emph{IEEE Transactions on
  Pattern Analysis and Machine Intelligence}, vol.~45, no.~4, pp. 5187--5202,
  2022.

\bibitem{yang2022robust}
M.~Yang, Y.~Li, P.~Hu, J.~Bai, J.~Lv, and X.~Peng, ``Robust multi-view
  clustering with incomplete information,'' \emph{IEEE Transactions on Pattern
  Analysis and Machine Intelligence}, vol.~45, no.~1, pp. 1055--1069, 2022.

\bibitem{trosten2023effects}
D.~J. Trosten, S.~L{\o}kse, R.~Jenssen, and M.~C. Kampffmeyer, ``On the effects
  of self-supervision and contrastive alignment in deep multi-view
  clustering,'' in \emph{Proceedings of the IEEE/CVF Conference on Computer
  Vision and Pattern Recognition}, 2023, pp. 23\,976--23\,985.

\bibitem{long2024s2mvtc}
Z.~Long, Q.~Wang, Y.~Ren, Y.~Liu, and C.~Zhu, ``S2mvtc: a simple yet efficient
  scalable multi-view tensor clustering,'' in \emph{Proceedings of the IEEE/CVF
  Conference on Computer Vision and Pattern Recognition}, 2024, pp.
  26\,213--26\,222.

\bibitem{liu2023contrastive}
J.~Liu, X.~Liu, Y.~Yang, Q.~Liao, and Y.~Xia, ``Contrastive multi-view kernel
  learning,'' \emph{IEEE Transactions on Pattern Analysis and Machine
  Intelligence}, vol.~45, no.~8, pp. 9552--9566, 2023.

\bibitem{wu2024low}
T.~Wu, S.~Feng, and J.~Yuan, ``Low-rank kernel tensor learning for incomplete
  multi-view clustering,'' in \emph{Proceedings of the AAAI Conference on
  Artificial Intelligence}, vol.~38, no.~14, 2024, pp. 15\,952--15\,960.

\bibitem{chen2023fast}
Z.~Chen, X.-J. Wu, T.~Xu, and J.~Kittler, ``Fast self-guided multi-view
  subspace clustering,'' \emph{IEEE Transactions on Image Processing}, 2023.

\bibitem{mi2024fast}
Y.~Mi, H.~Chen, Z.~Yuan, C.~Luo, S.-J. Horng, and T.~Li, ``Fast multi-view
  subspace clustering with balance anchors guidance,'' \emph{Pattern
  Recognition}, vol. 145, p. 109895, 2024.

\bibitem{zhong2023self}
G.~Zhong and C.-M. Pun, ``Self-taught multi-view spectral clustering,''
  \emph{Pattern Recognition}, vol. 138, p. 109349, 2023.

\bibitem{dornaika2024towards}
F.~Dornaika and S.~El~Hajjar, ``Towards a unified framework for graph-based
  multi-view clustering,'' \emph{Neural Networks}, vol. 173, p. 106197, 2024.

\bibitem{li2023robust}
C.~Li, H.~Che, M.-F. Leung, C.~Liu, and Z.~Yan, ``Robust multi-view
  non-negative matrix factorization with adaptive graph and diversity
  constraints,'' \emph{Information Sciences}, vol. 634, pp. 587--607, 2023.

\bibitem{dong2024centric}
Y.~Dong, H.~Che, M.-F. Leung, C.~Liu, and Z.~Yan, ``Centric graph regularized
  log-norm sparse non-negative matrix factorization for multi-view
  clustering,'' \emph{Signal Processing}, vol. 217, p. 109341, 2024.

\bibitem{liu2024sample}
S.~Liu, J.~Zhang, Y.~Wen, X.~Yang, S.~Wang, Y.~Zhang, E.~Zhu, C.~Tang, L.~Zhao,
  and X.~Liu, ``Sample-level cross-view similarity learning for incomplete
  multi-view clustering,'' in \emph{Proceedings of the AAAI Conference on
  Artificial Intelligence}, vol.~38, no.~12, 2024, pp. 14\,017--14\,025.

\bibitem{wan2024fast}
X.~Wan, B.~Xiao, X.~Liu, J.~Liu, W.~Liang, and E.~Zhu, ``Fast continual
  multi-view clustering with incomplete views,'' \emph{IEEE Transactions on
  Image Processing}, 2024.

\bibitem{zhang2015constrained}
X.~Zhang, L.~Zong, X.~Liu, and H.~Yu, ``Constrained nmf-based multi-view
  clustering on unmapped data,'' in \emph{Proceedings of the AAAI Conference on
  Artificial Intelligence}, vol.~29, no.~1, 2015, pp. 3174--3180.

\bibitem{wen2023unpaired}
Y.~Wen, S.~Wang, Q.~Liao, W.~Liang, K.~Liang, X.~Wan, and X.~Liu, ``Unpaired
  multi-view graph clustering with cross-view structure matching,'' \emph{IEEE
  Transactions on Neural Networks and Learning Systems}, pp. 1--15, 2023.

\bibitem{brbic2018multi}
M.~Brbi{\'c} and I.~Kopriva, ``Multi-view low-rank sparse subspace
  clustering,'' \emph{Pattern Recognition}, vol.~73, pp. 247--258, 2018.

\bibitem{zhang2023enhanced}
C.~Zhang, H.~Li, W.~Lv, Z.~Huang, Y.~Gao, and C.~Chen, ``Enhanced tensor
  low-rank and sparse representation recovery for incomplete multi-view
  clustering,'' in \emph{Proceedings of the AAAI Conference on Artificial
  Intelligence}, vol.~37, no.~9, 2023, pp. 11\,174--11\,182.

\bibitem{liu2012robust}
G.~Liu, Z.~Lin, S.~Yan, J.~Sun, Y.~Yu, and Y.~Ma, ``Robust recovery of subspace
  structures by low-rank representation,'' \emph{IEEE Transactions on Pattern
  Analysis and Machine Intelligence}, vol.~35, no.~1, pp. 171--184, 2012.

\bibitem{elhamifar2013sparse}
E.~Elhamifar and R.~Vidal, ``Sparse subspace clustering: Algorithm, theory, and
  applications,'' \emph{IEEE Transactions on Pattern Analysis and Machine
  Intelligence}, vol.~35, no.~11, pp. 2765--2781, 2013.

\bibitem{ng2001spectral}
A.~Ng, M.~Jordan, and Y.~Weiss, ``On spectral clustering: Analysis and an
  algorithm,'' \emph{Advances in Neural Information Processing Systems},
  vol.~14, 2001.

\bibitem{li2015large}
Y.~Li, F.~Nie, H.~Huang, and J.~Huang, ``Large-scale multi-view spectral
  clustering via bipartite graph,'' in \emph{Proceedings of the AAAI Conference
  on Artificial Intelligence}, vol.~29, no.~1, 2015.

\bibitem{kang2020large}
Z.~Kang, W.~Zhou, Z.~Zhao, J.~Shao, M.~Han, and Z.~Xu, ``Large-scale multi-view
  subspace clustering in linear time,'' in \emph{Proceedings of the AAAI
  Conference on Artificial Intelligence}, vol.~34, no.~04, 2020, pp.
  4412--4419.

\bibitem{wang2021fast}
S.~Wang, X.~Liu, X.~Zhu, P.~Zhang, Y.~Zhang, F.~Gao, and E.~Zhu, ``Fast
  parameter-free multi-view subspace clustering with consensus anchor
  guidance,'' \emph{IEEE Transactions on Image Processing}, vol.~31, pp.
  556--568, 2021.

\bibitem{wang2022align}
S.~Wang, X.~Liu, S.~Liu, J.~Jin, W.~Tu, X.~Zhu, and E.~Zhu, ``Align then
  fusion: Generalized large-scale multi-view clustering with anchor matching
  correspondences,'' \emph{Advances in Neural Information Processing Systems},
  vol.~35, pp. 5882--5895, 2022.

\bibitem{zhang2023let}
P.~Zhang, S.~Wang, L.~Li, C.~Zhang, X.~Liu, E.~Zhu, Z.~Liu, L.~Zhou, and
  L.~Luo, ``Let the data choose: Flexible and diverse anchor graph fusion for
  scalable multi-view clustering,'' in \emph{Proceedings of the AAAI Conference
  on Artificial Intelligence}, vol.~37, no.~9, 2023, pp. 11\,262--11\,269.

\bibitem{zhang2024learning}
C.~Zhang, X.~Jia, Z.~Li, C.~Chen, and H.~Li, ``Learning cluster-wise anchors
  for multi-view clustering,'' in \emph{Proceedings of the AAAI Conference on
  Artificial Intelligence}, vol.~38, no.~15, 2024, pp. 16\,696--16\,704.

\bibitem{10440580}
S.~Liu, Q.~Liao, S.~Wang, X.~Liu, and E.~Zhu, ``Robust and consistent anchor
  graph learning for multi-view clustering,'' \emph{IEEE Transactions on
  Knowledge and Data Engineering}, vol.~36, no.~8, pp. 4207--4219, 2024.

\bibitem{liu2024learn}
S.~Liu, K.~Liang, Z.~Dong, S.~Wang, X.~Yang, S.~Zhou, E.~Zhu, and X.~Liu,
  ``Learn from view correlation: An anchor enhancement strategy for multi-view
  clustering,'' in \emph{Proceedings of the IEEE/CVF Conference on Computer
  Vision and Pattern Recognition}, 2024, pp. 26\,151--26\,161.

\bibitem{huang2020partially}
Z.~Huang, P.~Hu, J.~T. Zhou, J.~Lv, and X.~Peng, ``Partially view-aligned
  clustering,'' \emph{Advances in Neural Information Processing Systems},
  vol.~33, pp. 2892--2902, 2020.

\bibitem{kuhn1955hungarian}
H.~W. Kuhn, ``The hungarian method for the assignment problem,'' \emph{Naval
  Research Logistics Quarterly}, vol.~2, no. 1-2, pp. 83--97, 1955.

\bibitem{munkres1957algorithms}
J.~Munkres, ``Algorithms for the assignment and transportation problems,''
  \emph{Journal of the Society for Industrial and Applied Mathematics}, vol.~5,
  no.~1, pp. 32--38, 1957.

\bibitem{yang2021partially}
M.~Yang, Y.~Li, Z.~Huang, Z.~Liu, P.~Hu, and X.~Peng, ``Partially view-aligned
  representation learning with noise-robust contrastive loss,'' in
  \emph{Proceedings of the IEEE/CVF Conference on Computer Vision and Pattern
  Recognition}, 2021, pp. 1134--1143.

\bibitem{serfozo2009basics}
R.~Serfozo, \emph{Basics of applied stochastic processes}.\hskip 1em plus 0.5em
  minus 0.4em\relax Springer Science \& Business Media, 2009.

\bibitem{lu2016fast}
Y.~Lu, K.~Huang, and C.-L. Liu, ``A fast projected fixed-point algorithm for
  large graph matching,'' \emph{Pattern Recognition}, vol.~60, pp. 971--982,
  2016.

\bibitem{huang2013spectral}
J.~Huang, F.~Nie, and H.~Huang, ``Spectral rotation versus k-means in spectral
  clustering,'' in \emph{Proceedings of the AAAI Conference on Artificial
  Intelligence}, vol.~27, no.~1, 2013, pp. 431--437.

\bibitem{wang2015projection}
W.~Wang and C.~Lu, ``Projection onto the capped simplex,'' \emph{arXiv Preprint
  arXiv:1503.01002}, 2015.

\bibitem{cai2013multi}
X.~Cai, F.~Nie, and H.~Huang, ``Multi-view k-means clustering on big data,'' in
  \emph{Twenty-Third International Joint Conference on Artificial
  Intelligence}, 2013, pp. 2598--2604.

\bibitem{cao2024view}
J.~Cao, W.~Dong, and J.~Chen, ``View-unaligned clustering with graph
  regularization,'' \emph{Pattern Recognition}, p. 110706, 2024.

\bibitem{glewis2004new}
Y.~Glewis, D.~David, and F.~Li, ``A new benchmark collection for text
  categorization research,'' \emph{J. Mach. Learn. Res}, vol.~15, pp. 361--397,
  2004.

\end{thebibliography}
\vfill
\end{document}